\documentclass[10pt,onecolumn,letterpaper]{article}
\usepackage{authblk}
\usepackage{iccv}
\usepackage{times}
\usepackage{epsfig}
\usepackage{graphicx}
\usepackage{makecell}
\usepackage{bm}
\usepackage{algorithm}
\usepackage{algpseudocode}
\usepackage{amsmath, amsthm, amsfonts, amssymb, color}

\usepackage[breaklinks=true,bookmarks=false]{hyperref}

\iccvfinalcopy 

\def\iccvPaperID{****} 

\ificcvfinal\fi

\begin{document}

\title{Compensation Learning}


\author[1,2]{Rujing Yao}
\author[2]{Ou Wu}
\affil[1]{Department of Information Resources Management, Nankai University, Tianjin, China}
\affil[2]{National Center for Applied Mathematics, Tianjin University, Tianjin, China}
\affil[1]{\textit {rjyao@mail.nankai.edu.cn}}
\affil[2]{\textit {wuou@tju.edu.cn}}

\maketitle
\ificcvfinal\thispagestyle{empty}\fi

{\bf Abstract---}Weighting strategy prevails in machine learning. For example, a common approach in robust machine learning is to exert lower weights on samples which are likely to be noisy or quite hard. This study reveals another undiscovered strategy, namely, compensating. Various incarnations of compensating have been utilized but it has not been explicitly revealed. Learning with compensating is called compensation learning and a systematic taxonomy is constructed for it in this study. In our taxonomy, compensation learning is divided on the basis of the compensation targets, directions, inference manners, and granularity levels. Many existing learning algorithms including some classical ones can be viewed or understood at least partially as compensation techniques. Furthermore, a family of new learning algorithms can be obtained by plugging the compensation learning into existing learning algorithms. Specifically, two concrete new learning algorithms are proposed for robust machine learning. Extensive experiments on image classification and text sentiment analysis verify the effectiveness of the two new algorithms. Compensation learning can also be used in other various learning scenarios, such as imbalance learning, clustering, regression, and so on.


\section{Introduction}

In supervised learning, a loss function is defined on the training set, and the training goal is to seek optimal models by minimizing the training loss. According to the degree of training difficulty, samples can be divided into easy, medium, hard, and noisy samples. Generally, easy and medium samples are indispensable and positively influence the training. The whole training procedure can significantly benefit from medium samples if appropriate learning manners are leveraged. However, the whole training procedure is vulnerable to noisy and partial quite hard samples.

A common practice is to introduce the weighting strategy if hard and noisy samples exist. Low weights are assigned to noisy and quite hard samples to reduce their negative influences during loss minimization. This strategy usually infers the weights and subsequently conducts training on the basis of the weighted loss~\cite{liu2015classification}. Wang et al.~\cite{wang2017robust} proposed a Bayesian method to infer the sample weights as latent variables. Kumar et al.~\cite{kumar2010self} proposed a self-paced learning (SPL) manner that combines the two steps as a whole by using an added regularizer. Meta learning~\cite{li2019learning,ren2018learning,wang2020training} is introduced to alternately infer weights and seek model parameters with an additional validation set.

Various robust learning methods exist that do not rely on the weighting strategy. For example, the classical method support vector machine (SVM)~\cite{cortes1995support} introduces slack variables to address possibly noisy samples, and robust clustering~\cite{forero2012robust} introduces additional vectors to cope with noises. However, a unified theory to better explain such methods and subsequently illuminate more novel methods remains lacking. In this study, another under-explored yet widely used strategy, namely, compensating, is revealed and further investigated. Mathematically, the compensating strategy actually adds\footnote{Weighting actually multiplies a term to a feature vector, a logit vector, a loss, etc.} a perturbation term (called compensation term in this study) to a feature vector, a logit vector, a loss, etc. Many existing learning methods including some classical ones can be considered introducing or partial on the basis of compensating. Learning with compensating is referred to as compensation learning in this study.

We conduct a pilot study for compensation learning in terms of theoretical taxonomy, connections with existing classical learning methods, and new concrete compensation learning methods. First, five compensation targets, three directions, five inference manners, and four granularity levels are defined. Second, several existing learning methods are re-explained from the viewpoint of compensation learning. Third, two concrete compensation learning algorithms are proposed, namely, logit compensation with $l$1-regularization (LogComp) and mixed  compensation (MixComp). Last, the two proposed learning algorithms are evaluated on data corpora from image classification and text sentiment classification.

Our main contributions are summarized as follows:

1) An under-explored yet widely used learning strategy, namely, compensating, is identified and formalized in this study. A new learning paradigm, named compensation learning, is presented and a taxonomy is constructed for it. In addition to the robust learning mainly referred in this paper, other learning scenarios, such as imbalance learning can also benefit from compensation learning.

2) Several typical learning methods are re-explained with the viewpoint of compensation learning. New insights can be obtained for these methods. Theoretically, various new methods can be generated on the basis of introducing the idea of compensating into existing methods. Section V and VI-D present examples.

3) Two concrete new compensation learning methods are proposed. Experiments on robust learning on four benchmark sets verify their effectiveness compared with several existing classical methods.

\section{Related Work}
\subsection{The Weighting Strategy in Machine Learning}
Weighting is a widely used machine learning strategy in at least the following five areas: noise-aware learning~\cite{natarajan2013learning}, curriculum learning~\cite{bengio2009curriculum}, crowdsourcing learning~\cite{deng2013fine}, cost-sensitive learning~\cite{chang2011ordinal}, and imbalance learning~\cite{huang2016learning}. In noisy-aware and curriculum learning areas, weights are sample-wise; in cost-sensitive learning, weights can be sample-wise, category-wise, or mixed; in imbalance learning, weights are usually category-wise.

Intuitively, the weights of medium and partial hard samples are kept or enlarged; and the weights of quite hard samples should be kept or reduced. For example, in Focal loss~\cite{lin2017focal}, the weights of easy samples are (relatively) reduced and those of the hard\footnote{In fact, if the weights of quite hard samples are reduced, the performance will be increased~\cite{li2019gradient}.}  samples are (relatively) enlarged. Most existing studies do not assume the above division. Instead, samples are usually divided into easy/non-easy or normal/noisy. For example, in Focal loss and Adaboost~\cite{freund1997decision}, the weights of non-easy samples are gradually increased.

In cost-sensitive learning, the weights are associated with the pre-determined costs. In imbalance learning, categories with lower proportions are usually negatively affected. Therefore, increasing the weights of samples in the low-proportion categories is a common practice.

The compensating strategy investigated in this study does not intend to eliminate the weighting strategy. Instead, this study summarizes various existing learning ideas which do not utilize weighting yet. These learning ideas are systematically investigated to attribute to a new learning paradigm, namely, compensation learning. These two strategies can be mutually beneficial\footnote{For example, a sample-level weighting method (e.g., Focal loss) can be transformed into a category-level weighting method (e.g., replace the sample-level prediction $y_i$ with the category-level average $y_c$) inspired by our taxonomy for compensating learning.}. Theoretically, each concrete weighting-based learning method may correspond to a concrete compensating-based learning method. A solid and deep investigation for the weighting strategy in machine learning will significantly benefit compensation learning.

\subsection{Noise-aware Machine Learning}\label{sec:sec22}
This study investigates compensation learning mainly in learning with noisy labels. The weighting strategy is prevailing in this area. There exist two common technical solutions.

In the first solution, noise detection is performed and noisy samples may be assigned lower weights in the successive model training. Koh and Liang~\cite{han2018co} defined an influence function to measure the impact of each sample on the model training. Samples with higher influence scores are more likely to be noisy. Huang et al.~\cite{huang2019o2u} conducted a cyclical pre-training strategy and recorded the training losses for each sample in the whole cycles. The samples with higher average training losses are more likely to be noisy.

In the second solution, an end-to-end procedure is leveraged to construct a noise-robust model. Reed et al.~\cite{reed2014training} proposed a Bootstrapping loss to reduce the negative impact of samples which may be noisy. Goldberger and en-Reuven~\cite{goldberger2016training} designed a noise adaptation layer to model the relationship between overserved labels that may be noisy and true latent labels.

A recent survey can be referred to~\cite{han2020survey}. Compensation learning can replace weighting in both above solutions. In this study, only the second solution is referred.

\subsection{Robust Machine Learning}
A formal definition for robust machine learning does not exist at present. There are two typical learning scenarios for robust machine learning. The first scenario refers to the robustness of a learning process, while the second scenario refers to the robustness of a trained model. In the first scenario, a robust learning method should cope well with training data that may be noisy~\cite{song2020learning,Yao2019Deep}, imbalance~\cite{johnson2019survey}, few-shot~\cite{wang2020generalizing,Das2019A}, etc. In the second scenario, a robust trained model should cope well with adversarial attacks~\cite{zhang2020adversarial}. Both scenarios receive much and increasing attention in recent years. Both the weighting and the compensating strategies are widely-used in the first scenario, whereas only the compensating strategy is mainly utilized in the second scenario.

\begin{figure}[t]
    \centering
    \includegraphics[width=0.5\linewidth]{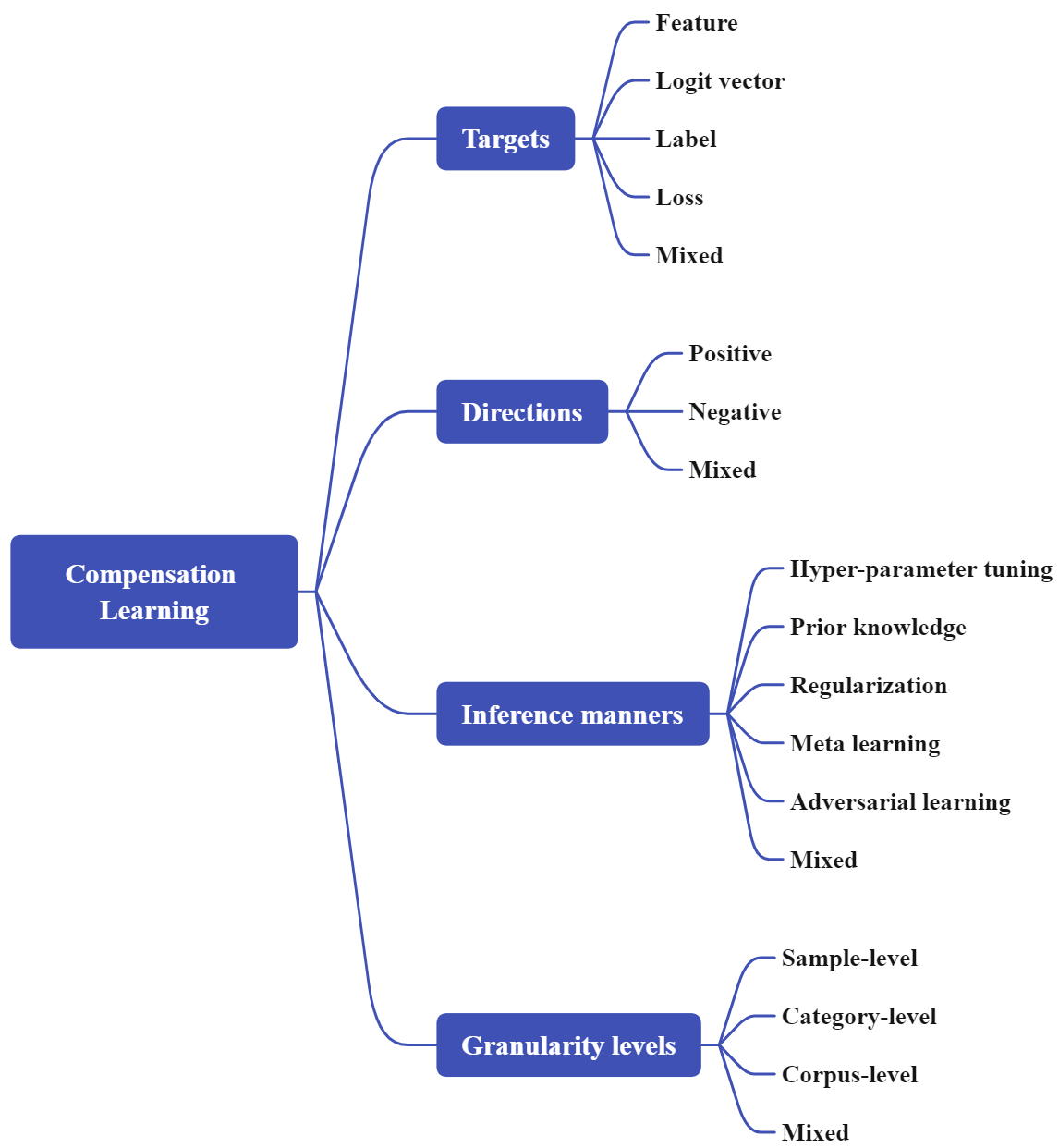}
    \caption{Taxonomy of compensation learning.}
    \label{fig:fig0}
\end{figure}

\section{A Taxonomy of Compensation Learning}\label{sec:sec3}
Compensating can be used in many learning scenarios. This section leverages classification as the illustrative example. Given a training set $S = \{x_i, y_i\}$, $i = 1, \ldots, N$, where $x_i$ is the $i$-th sample, and $y_i \in \{1, \ldots, c, \ldots, C\}$ is its categorical label. In a standard supervised deep learning context, let $u_i$ be the logit vector for $x_i$ output using a deep neural network. The training loss can be written as follows:
\begin{equation}\label{1-1}
\begin{aligned}
\mathcal{\mathcal{L}}  &= \sum\nolimits_i{l(\mathbb{S}({f}({x_i})),{y_i})}  = \sum\nolimits_i{l(\mathbb{S}({u_i}),{y_i})},  
\end{aligned} 
\end{equation}
where $\mathbb{S}(\cdot)$ transforms the logit vector $u_i$ into a soft label $p_i$, ${f(\cdot)}$ represents a deep neural network, and $u_i = f(x_i)$.

In the weighting strategy, the loss function is usually defined as follows:
\begin{equation}\label{2}
\mathcal{L}  = \sum\nolimits_i{{w_i \cdot}l(\mathbb{S}({u_i}),{y_i})},
\end{equation}
where $w_i$ is the weight associated to the sample $x_i$. Theoretically, the more likely a sample is noisy or quite hard, the lower its weight.

The compensating strategy investigated in this study can also increase or reduce the influences of samples in model training on the basis of their degrees of training difficulty. For instance, a negative value can be added to reduce the loss incurred from a noisy sample. Resultantly, the negatively influence of this sample will be reduced because its impact on gradients is reduced. Contrarily, when the influence should be increased, a positive value can be added to the loss incurred from the sample. In terms of mathematical computation, ``weighting" relies on the multiplication operation, whereas “compensating” relies on adding operation.

Fig.~\ref{fig:fig0} shows the constructed taxonomy of the whole compensation learning. This section introduces each item in the taxonomy.

\subsection{Compensation Targets}

Eq.~(\ref{1-1}) contains four different types of variables for each sample, namely, raw feature $x_i$, logit vector $u_i$, label $y_i$, and sample loss $l_i$~(=$l(\mathbb{S}({u_i}),{y_i})$). Therefore, compensation targets can be feature, logit vector, label, and loss.

(1) {\bf Compensation for feature (Feature compensation)}. In this kind of compensation, the raw feature vector ($x_i$) or transformed feature vector (e.g., dense feature) of each sample can have a compensation vector ($\Delta{x_i}$). Eq.~(\ref{1-1}) becomes
\begin{equation}\label{3}
\begin{aligned}
\mathcal{L}  &= \sum\nolimits_i {l(\mathbb{S}(f({x_i}{\rm{ + }}\Delta {x_i})),{y_i})} = \sum\nolimits_i {l(\mathbb{S}({u_i'}),{y_i})}.
\end{aligned} 
\end{equation}

(2) {\bf Compensation for logit vector (Logit compensation)}. In this kind of compensation, the logit vector ($u_i$) of each sample can have a compensation vector ($\Delta{u_i}$). Eq.~(\ref{1-1}) becomes
\begin{equation}\label{4}
\mathcal{L}  = \sum\nolimits_i {l(\mathbb{S}({u_i}{\rm{ + }}\Delta {u_i}),{y_i})}.
\end{equation}

(3) {\bf Compensation for label (Label compensation)}. In this kind of compensation, the label ($y_i$) of each sample can have a compensation label ($\Delta y_i$). Let $p_i = \text{softmax}(u_i)$. Eq.~(\ref{1-1}) becomes
\begin{equation}\label{5}
\begin{array}{l}
({\rm{i}}) \quad {\rm{ }}\mathcal{L}  = \sum\nolimits_i {l({p_i},{y_i}{\rm{ + }}\Delta {y_i})} {\quad\rm{    or}} \vspace{1mm}\\ 
({\rm{ii}})\quad {\rm{  }}\mathcal{L}  = \sum\nolimits_i {l({p_i}{\rm{ + }}\Delta {y_i},{y_i})}.
 \end{array}
\end{equation}

In Eq.~(\ref{5}-i), $\Delta y_i$ is added to the true label $y_i$, while in (ii) $\Delta y_i$ is added to the predicted label $p_i$. Considering that labels after compensation should be a (soft) label, $\Delta y_i$ should satisfy the following requirements:  
\begin{equation}\label{6}
\sum\nolimits_c {\Delta {y_{ic}} = 0, {y_{ic}} + \Delta {y_{ic}}{\rm{ }} \ge {\rm{ 0 \quad or \quad }}{p_{ic}} + \Delta {y_{ic}}{\rm{ }} \ge {\rm{ 0}}}.
\end{equation}

(4) {\bf Compensation for loss (Loss compensation)}. In this kind of compensation, the loss of each sample can have a compensation loss $(\Delta l_i)$. Eq.~(\ref{1-1}) becomes
\begin{equation}\label{7}
\mathcal{L}  = \sum\nolimits_i {l(\mathbb{S}({u_i}),{y_i}){\rm{ + }}\Delta } {l_i}.
\end{equation}

(5) {\bf Compensation for mixed targets (Mix-target compensation)}. In this kind of compensation, two or more of the aforementioned targets can have their compensation terms, simultaneously. For example, when both feature and label compensations are utilized, Eq. (3) becomes
\begin{equation}\label{7-1}
\begin{aligned}
\mathcal{L}  &= \sum\nolimits_i {l(\mathbb{S}(f({x_i}{\rm{ + }}\Delta {x_i})),{y_i}+\Delta {y_i})},
\end{aligned} 
\end{equation}
where $\Delta {x_i}$ and $\Delta {y_i}$ are the feature and loss compensations, respectively\footnote{Lee et al.~\cite{lee2020semantics} combine adversarial training and label smoothing, which can be considered as mix-target (feature and label) compensation.}.

{\bf Remark:} The compensation variables (i.e., $\Delta x_i$, $\Delta u_i$, $\Delta y_i$, and $\Delta l_i$) are trainable during training. They are introduced to reduce the negative impact of some training samples (e.g., noisy or partial hard ones) and increase the positive impact of some samples (e.g., medium ones). For example, in raw feature-based compensation, let $m_{y_i}$ be the center vector of the category of $x_i$. Ideally, if $\Delta x_i$ = $m_{y_i}$ – $x_i$, the impact of $x_i$ is completely reduced if $x_i$ is noisy.

If the loss functions defined in Eqs.~(\ref{3})--(\ref{7-1}) are directly used without any other constrictions on the compensation variables, nothing can be learned as compensation variables are trainable. For example, when the loss in Eq.~(\ref{4}) is directly used, a random model will be produced because in the training, the value of $\Delta u_i$ will be learned to be equal to $y_i$. How to infer them and learn with the above loss functions are described in the succeeding subsection.

There may exist other compensation candidates, such as view, structure (e.g., adjacency matrix in GCN), and gradient, which will be explored in future work.

\subsection{Compensation Directions}
 There are two directions according to the loss variations after compensation.

(1) {\bf Positive compensation}. If the compensation reduces the loss, then it is called positive compensation. Positive compensation can reduce the influence of noisy and quite hard samples during training.

(2) {\bf Negative compensation}. If the compensation increases the loss, then it is called negative compensation. Negative compensation can increase the influence of easy and medium samples during training. This case will be discussed in the rest of this paper.

(3) {\bf Mix-direction compensation}. If the compensation increases the losses of some training samples and decreases the losses of others simultaneously, then it is called mix-direction compensation. The logit adjustment method actually leverages this type of compensation, which will be discussed in Section IV.

\subsection{Compensation Inference}
In compensation learning, compensation variables in losses in Eqs.~(\ref{3})--(\ref{7}) should be inferred during training. There are five typical manners (maybe not exhaustive) to infer their values and optimize the whole loss.

(1) {\bf Inference with prior knowledge}.
In this manner, the compensation variables are inferred on the basis of prior knowledge. Alternatively, the compensation variables are fixed before the optimizing of training loss. Taking the label compensation as an example. Given that for each sample, we can obtain a predicted label $y'_i$ by another model, the label compensation can be defined as 
\begin{equation}\label{12}
\Delta {y_i}{\rm{ = }}\lambda  ({y'_i} - {y_i}),
\end{equation}
where $\lambda$ is a hyper-parameter and locates in [0, 1]. $\Delta y_i$  defined in Eq.~(\ref{12}) satisfies the condition given by Eq.~(\ref{6}). If $y'_i$ is in trust, then it is highly possible that $\Delta y_i$  approaches to zero if $x_i$ is normal, and it is large if $x_i$ is noisy. Assuming that $y'_i$ is the output of the model in the previous epoch. Eq.~(\ref{5}-i) becomes
\begin{equation}\label{13}
\mathcal{L} = \sum\nolimits_i{l(\mathbb{S}({u_i}),{y_i}{\rm{ + }}\lambda ({y'_i} - {y_i}))},
\end{equation}
which is exactly the Bootstrapping loss~\cite{reed2014training}.

(2) {\bf Inference with hyper-parameter tuning}. In this manner, the compensation variable(s) is/are taken as hyper-parameter(s). Consequently, the optimal value is determined according to the manner of hyper-parameter tuning.

(3) {\bf Inference with regularization}. In this manner, a regularization term is added for the compensation variables. For example, a natural assumption is that the proportion of the samples that require the compensation variables is small. Therefore, $l1$-norm can be used. Taking the logit compensation as examples. A loss function is defined as follows:
\begin{equation}\label{8}
 {\rm{  }}\mathcal{L}  = \sum\nolimits_i {l(\mathbb{S}({u_i} + \Delta {u_i}),{y_i})}  + \lambda {\mathop{ Reg}\nolimits} {\rm{(}}\Delta {u_i}), 
\end{equation}
where $\lambda$ is a hyper-parameter and $Reg(\cdot)$ is regularizer. This manner is similar to the self-paced learning~\cite{kumar2010self}. When $\lambda \to \infty$, no compensation is allowed and compensation learning is reduced to conventional learning.

(4) {\bf Inference with meta learning}. In this manner, the compensation variables are inferred on the basis of another small clean validation set with meta learning. Given a clean validation set $\Omega$ comprising $M$ clean training samples and taking loss compensation as an example. Let $\kappa _i$ be the loss compensation variable for $x_i$ $(\in S)$. We first define that 
\begin{equation}\label{9}
\mathcal{L} = \sum\nolimits_{i \in S} {l(\mathbb{S}({u_i}),{y_i}:\Theta )} {\rm{ + }}{\kappa _i},
\end{equation}
where $\Theta$ is the model parameter set to be learned. Given $\bm{\kappa}$, $\Theta$ can be optimized on the training set $S$ by solving
\begin{equation}\label{10}
{\Theta ^{\rm{*}}}(\bm{\kappa} ) = \arg \mathop {\min }\limits_\Theta  \sum\nolimits_{i \in S} {l(\mathbb{S}({u_i}),{y_i}:\Theta )} {\rm{ + }}{\kappa _i}.
\end{equation}

After $\Theta$ is obtained, $\bm{\kappa}$ can be optimized on the validation set $\Omega$ by solving
\begin{equation}\label{11}
{\bm{\kappa} ^{\rm{*}}} = \arg \mathop {\min }\limits\bm{_\kappa} \sum\nolimits_{j \in \Omega} {l(\mathbb{S}({u_j}),{y_j}:{\Theta ^*}(\bm{\kappa} ))}.
\end{equation}

These two optimizations can be performed alternately, and finally $\Theta^{*}$ and $\bm{\kappa^{*}}$ are learned. When either logit or label compensation is used, the above optimization procedure can also be utilized with slight variations.

The above inference manner is similar with that used in the meta learning-based weighting strategy for robust learning~\cite{ren2018learning}. Meta learning has been widely used in robust learning and many existing meta learning-based weighting methods~\cite{li2019learning, wang2020training} can be leveraged for compensation learning.

(5) {\bf Inference with adversarial learning}. In both feature and logit compensations, the compensation term can be obtained by adversarial learning. Taking feature compensation as an example, the objective function in negative compensation is 
\begin{equation}\label{3-1}
\begin{aligned}
\Delta {x_i^{*}} &= \arg \mathop {\max }_{{\left\| {\Delta {x_i}} \right\| \le \epsilon}} {l(\mathbb{S}(f({x_i}{\rm{ + }}\Delta {x_i})),{y_i})}, 
\end{aligned} 
\end{equation}
where $\epsilon$ is the bound. Likewise, the objective function in positive feature-level compensation can be
\begin{equation}\label{3-2}
\begin{aligned}
\Delta {x_i^{*}} &= \arg \mathop {\min }_{{\left\| {\Delta {x_i}} \right\| \le \epsilon}} {l(\mathbb{S}(f({x_i}{\rm{ + }}\Delta {x_i})),{y_i})}. 
\end{aligned} 
\end{equation}

(6) {\bf Inference with mixed manners}. Two or more of the above five manners can be combined together to infer the compensation term in a learning task.

{\bf Remark:} An existing compensation-based learning method usually adopts one of the inference manners listed above. Theoretically, the inference manner can be changed from one manner to another and a new method will subsequently be obtained. 

\subsection{Compensation Granularity}

Compensation granularity has four levels.

(1) {\bf Sample-level compensation}. All the compensation variables discussed above are for samples. Each sample has its own compensation variable.

(2) {\bf Category-level compensation}. In this level, samples within the same category share the same compensation. Taking the logit vector-based compensation as an example, when category-level compensation is utilized, the loss in Eq.~(\ref{4}) becomes
\begin{equation}\label{14}
\mathcal{L}  = \sum\nolimits_i {l(\mathbb{S}({u_i}{\rm{ + }}\Delta {u_{{y_i}}}),{y_i})}.
\end{equation}

Category-level compensation mainly solves the problem when the impact of all the samples of a category should be increased. For example, in long-tail classification, the tail category should be emphasized in learning.

(3) {\bf Corpus-level compensation}. 
In this level, samples within the whole training corpus share the same compensation. Take the negative compensation described in Eq.~(\ref{3-1}) as an example, the objective function becomes
\begin{equation}\label{3-3}
\begin{aligned}
\Delta {x^{*}} &= \arg \mathop {\max }_{{\left\| {\Delta {x}} \right\| \le \epsilon}} {l(\mathbb{S}(f({x_i}{\rm{ + }}\Delta {x})),{y_i})}, 
\end{aligned} 
\end{equation}
which means that all samples share the same term $\Delta {x^{*}}$. $\Delta {x^{*}}$ is exactly the universal adversarial perturbation~\cite{Moosav2020}.

(4) 
{\bf Mix-level compensation}. In this level, more than one of the aforementioned three levels are utilized simultaneously. This case occurs in complex contexts, e.g., when noisy labels and category imbalance exist. Taking label-based compensation as an example. The loss in Eq.~(\ref{4}) can be written as\vspace{-2mm}
\begin{equation}\label{15}
\vspace{-2mm}
{\rm{ }}\mathcal{L}  = \sum\nolimits_i {l({p_i},{y_i}{\rm{ + }}\Delta {y_i}{\rm{ + }}\Delta {y_{{y_i}}})},
\end{equation}
where $\Delta {y_{{y_i}}}$ is the category-level label compensation.


\section{Connection with Existing Learning Paradigms}

The weighting strategy is straightforward and quite intuitive, hence it has been widely used in the machine learning community. Compensating seems not as straightforward as weighting. However, it can play the same/similar role with weighting in machine learning. They both have their own strengths. Compensating can be used in the feature, logit vector, label, and loss, whereas weighting is usually used in the loss. Weighting is usually efficient, whereas the optimization in some compensating methods (e.g., Eq.~(\ref{3-3})) is relatively complex. A theoretical comparison for them is beneficial for both strategies and we leave it as future work.

Many classical and newly proposed learning methods, which are on the basis of distinct inspirations and theoretical motivations, can be attributed to compensation learning or explained in the viewpoint of compensation learning. We choose the following methods as illustrative examples.

(1) {\bf Robust clustering}~\cite{forero2012robust}. Let $m_c$ be the cluster center of the $c$-th cluster. Let $\omega_{ic}$ ($\in \{0, 1\}$) denote whether $x_i$ belongs to the $c$-th cluster. The optimization form of conventional data clustering can be written as follows:
\begin{equation}\label{20}
\mathop {\min }\limits_{\{ {m_c}\} ,\{ {\omega_{ic}}\} } \sum\nolimits_i {\sum\nolimits_c {{\omega_{ic}}} } \left\| {{x_i} - {m_c}} \right\|_2^2.
\end{equation}

Given that outlier samples may exist, sample-level feature compensation (denoted as $o_i$ for $x_i$) can be introduced with regularization.  When $l2$-norm is used,~(\ref{20}) becomes
\begin{equation}\label{22}
\mathop {\min }\limits_{\{ {m_c}\} ,\{ {\omega_{ic}}\}, {\rm{\{ }}{o_i}{\rm{\} }}} \sum\nolimits_i{\sum\nolimits_c {{\omega_{ic}}\left( {\left\| {({x_i} + {o_i}) - {m_c}} \right\|_2^2 + \lambda {{\left\| {{o_i}} \right\|}_2}} \right)} },
\end{equation}
which becomes the method proposed by Foreo et al.~\cite{forero2012robust}.

(2) {\bf Adversarial training}. An adversarial sample can be regarded as a negatively compensated one for the original sample. Training with adversarial samples (i.e., adversarial training) is proven to be useful in many applications and various methods are proposed~\cite{madry2017towards}.

Shafahi et al.~\cite{shafahi2020universal} proposed universal adversarial training which is actually based on a corpus-level negative feature compensation. The loss on adversarial samples is
\begin{equation}
\mathcal{L}_{corpus-adv} = \mathop {\max }\limits_{{{\left\| \delta  \right\|}} \le \epsilon } \sum\nolimits_i {l\left( {\mathbb{S}(f({x_i} + \delta)),y_i } \right)}. \end{equation}

Benz et al.~\cite{benz2021universal} observed that universal adversarial perturbation does not attack all classes equally. They proposed a category-wise universal adversarial training method and the loss on adversarial samples is
\begin{equation}
\mathcal{L}_{category-adv} = \mathop {\max }\limits_{{{\left\| \delta_{y_i}  \right\|}} \le \epsilon } \sum\nolimits_i {l\left( {\mathbb{S}(f({x_i} + \delta_{y_i})),y_i } \right)}, \end{equation}
which belongs to the category-level negative feature compensation. Motivated by our taxonomy, mixed corpus/category/sample-level adversarial perturbations can subsequently be generated. A mixed corpus/sample-level adversarial perturbation is described as an example:
\begin{equation}\label{ss-1}
\begin{array}{l}
\delta^* = arg\underset{\delta}{\max} \sum\limits_i {l(S(f(x_i + \delta )),{y_i})}  \\ 
 {\mathcal{L}_{mix-adv}} = \underset{\delta_i}\max \sum\limits_i {l(S(f(x_i + \delta^*  + {\delta _i})),{y_i})},  \\ 
 \end{array}
\end{equation}
where $\delta^*$ and $\delta_{i}$ are the corpus-level and sample-level perturbations, respectively. A further statistical analysis for the two levels of adversarial perturbations may illuminate us to better understand the adversarial characteristics of the data.

(3) {\bf Adversarial label smoothing~\cite{goibert2019adversarial}}. Label smoothing is actually a type of sample-level label compensation. Its compensation term for a sample ($x_i, y_i$) is defined as follows:
\begin{equation}\label{25-1}
\Delta{y_i}  = \lambda(I/C-y_i),
\end{equation}
where $I$ is a $C$-dimensional vector and each element is equal to 1. Obviously, the compensation term is determined according to pre-definition (i.e., the prior knowledge manner). 

According to the inference manner in our taxonomy, adversarial learning can be utilized to pursue the compensation term. Accordingly, the term is 

\begin{equation}\label{25-2}
\begin{aligned}
\Delta{y_i} = \lambda(p^*_i-y_i),
\end{aligned}
\end{equation}
\text{where} \\
\begin{equation}\label{25-3}
\begin{aligned}
p^*_i = arg\underset{p_i}{\max} {l(\mathbb{S}(u_i),y_i+\lambda(p_i-y_i))}.
\end{aligned}
\end{equation}
Eq.~(27) has an analytic solution such that $p^*_i$ is the one-hot vector for the category which corresponds to the minimum softmax value in $\mathbb{S}(u_i)$.

(4) {\bf Logit adjustment-based imbalance learning}~\cite{menon2020long}. In a multi-category classification problem, let $\pi _c$ be the proportion of the training samples in the $c$-th category. Let $\textbf{g} = \left[ {g\left( {{\pi _1}} \right), \ldots, g\left( {{\pi _C}} \right)} \right]$. When the proportions are imbalanced, a corpus-level of logit compensation can be introduced as follows:
\begin{equation}\label{25}
\mathcal{L}  = \sum\nolimits_i {l\left( {\mathbb{S}({u_i}{\rm{ + }}\textbf{g}),{y_i}} \right)}.
\end{equation}

For the above loss, when ${g}(\cdot)$ is an increasing function, we conjecture that the influences of samples in the minority categories (i.e., ${\pi _c} < \frac{1}{C}$) on the loss are increased. 
\begin{figure}[t]
    \centering
    \includegraphics[width=0.6\linewidth]{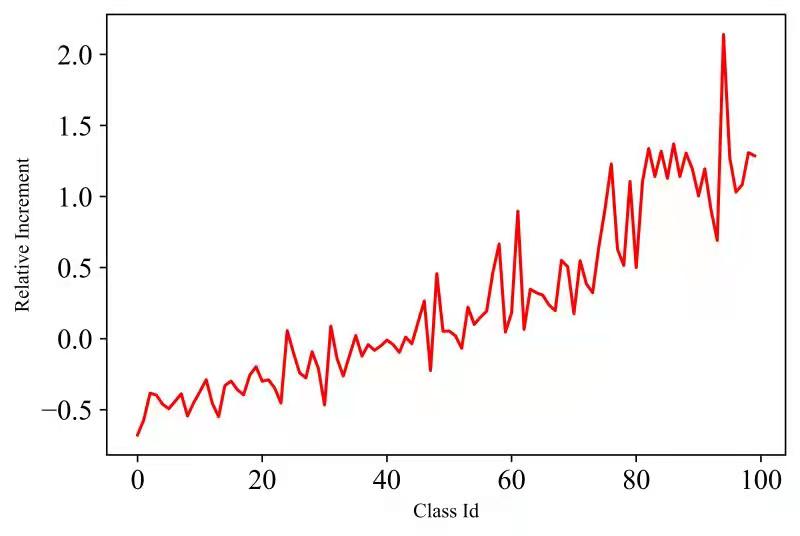}
    \caption{The relative loss increment ($(l'-l)/{l}$) for Logit Adjustment. Head categories are in the left and tail ones are in the right. The losses of head categories are mainly decreased, while those of tail ones are increased.}
    \label{fig:figa1}
\end{figure}

As the influences of samples in the minority categories on the loss are increased, the imbalanced problem can be alleviated by the logit compensation used in Eq.~(\ref{25}). When ${g}\left( {{\pi _c}} \right) = \tau \log \left( {{\pi _c}} \right)\left( {\tau  > 0} \right)$ and cross-entropy loss are used, Eq.~(\ref{25}) becomes

\begin{equation}\label{pp26}
\mathcal{L}  = {\rm{ - }}\sum\nolimits_i {\log \frac{{{e^{{u_{i,{y_i}}} + \tau \log {\pi _{{y_i}}}}}}}{{\sum\nolimits_c {{e^{{u_{i,c}} + \tau \log {\pi _c}}}} }}},
\end{equation}
which is exactly the logit adjusted loss~\cite{menon2020long}. We make a statistic for the relative loss variations incurred by Logit adjustment for each category on the imbalanced version of the benchmark image classification data CIFAR-100~\cite{krizhevsky2009learning}. The results are presented in Fig.~\ref{fig:figa1}. The
loss variations of head categories (those with small class Ids) are negative, and those of tail (those with small class Ids) are positive. In other words, both positive  and negative compensations exist in logit adjustment. Intuitively, a category-level version can be obtained via meta learning, which is discussed in Appendix~\ref{app:4}.

(5) {\bf SVM~\cite{cortes1995support}.} This method is based on the following hinge loss:
\begin{equation}\label{a16}
{\rm{ }}{l_i} = \max (0,{\rm{ }}1 - {y_i}({\text{w}^T}{x_i} + b)).
\end{equation}
To reduce the negative contributions of noisy or hard samples, the loss can be compensated as follows:
\begin{equation}\label{a17}
\begin{aligned}
 {\rm{ }}{{l'}_i} &= \max (0,{\rm{ }}{l_i}{\rm{ - }}{\xi _i})\quad{\rm{   }}   \\ 
 &= \max (0, 1 - {y_i}({\rm{\textbf{w}}^T}{x_i} + b){\rm{ - }}{\xi _i}){\rm{ }} {\qquad\rm{(}}  {\xi _i} \ge 0{\rm{)}}. 
\end{aligned}  
\end{equation}
Then the whole training loss with max margin and $l1$-norm for ${\xi _i}$ becomes
\begin{equation}\label{a18}
{\rm{ }}\mathcal{L}  = \frac{1}{2}{\left\| {\rm{\textbf{w}}} \right\|^2} + \sum\nolimits_i {{{l'}_i} + } \lambda {\rm{|}}{\xi _i}{\rm{|\qquad   (}}{\xi _i} \ge 0{\rm{)}}.
\end{equation}
The minimization of Eq.~(\ref{a18}) equals to the following optimization problem:
\begin{equation}\label{a19}
\begin{array}{l}
 {\min\limits_{\textbf{w},b,\{\xi _i\}}  \frac{1}{2}{\left\| {\rm{\textbf{w}}} \right\|^2} + \lambda \sum\nolimits_i {{\xi _i}} {\rm{ }}} \\ 
 {\rm{s}}{\rm{.t}}{\rm{.  }} \quad 1 - {y_i}({{\textbf{w}^T}{x_i} + b){\rm{ - }}{\xi _i}} \le 0\\
 \qquad{\rm{  }}{\xi _i} \ge 0, i=1,\cdots,N \\ 
 \end{array},
\end{equation}
which is the standard form of SVM (without kernel). Alternatively, the slack variable can be seen as a loss compensation for SVM. Naturally, other types of compensation (e.g., label compensation) may be considered in SVM.

(6) {\bf Knowledge distillation~\cite{hinton2015distilling}.} In knowledge distillation, there are two deep neural networks called teacher and student, respectively. The output of the teacher model for $x_i$ is
\begin{equation}\label{23}
{q_i} = \text{\text{softmax}}({z_i}/T),
\end{equation}
where $z_i$ is the logit vector from the teacher model and $T$ is the temperature. $q_i$ can be viewed as a prior knowledge about the label compensation for the student model. 

Then according to Eq.~(\ref{6}), the training loss of the student model with label compensation becomes
\begin{equation}\label{aaa24}
\begin{aligned}
 \mathcal{L}  &= \sum\nolimits_i {l({p_i},{y_i}{\rm{ + }}\lambda ({q_i} - {p'_i}))},
\end{aligned} 
\end{equation}
where $p'_i = \text{\text{softmax}}({u_i}/T)$. Eq.~(\ref{aaa24}) is exactly the loss function of knowledge distillation.

(7) {\bf Implicit semantic data augmentation (ISDA) ~\cite{wang2019implicit}.} In contrast with previous data augmentation techniques, ISDA does not produce new samples or features. Instead, ISDA transforms the semantic data augmentation problem into the optimization of a new loss defined as
\begin{equation}
\small
\mathcal{L}{\rm{ =  - }}\sum\nolimits_{i } {\frac{{\exp ({u_{i,{y_i}}})}}{{\sum\limits_{c = 1}^C {\exp ({u_{i,{c}}} + \frac{\lambda }{2}{{({{\rm{w}}_c} - {{\rm{w}}_{{y_i}}})}^T}{\Sigma _{{y_i}}}({{\rm{w}}_c} - {{\rm{w}}_{{y_i}}}))} }}},
\end{equation}
where $\Sigma_{y_i}$ is is the co-variance matrix for the $y_i$th category; $\text{w}_c$ is the model parameter for the logit vectors and ${u_{i,c}  = }{{\rm{w}}_{\rm{c}}}^T{\tilde x_i}{{ }}$ ($\tilde x_i$ is the output of the last feature encoding layer for the sample $x_i$).

In Eq. (36), a logit compensation term is observed as follows:
\begin{equation}
\begin{aligned}
u'_i=u_i+\delta_{y_i},\\
\end{aligned}
\end{equation}
\text{where}\\
\begin{equation}
\begin{aligned}
{\delta _{{y_i}}}{\rm{ = }}\frac{\lambda }{2}\left[ {\begin{array}{*{20}{c}}
{{{({{\rm{w}}_1} - {{\rm{w}}_{{y_i}}})}^T}{\Sigma _{{y_i}}}({{\rm{w}}_1} - {{\rm{w}}_{{y_i}}})}\\
 \vdots \\
{{{({{\rm{w}}_C} - {{\rm{w}}_{{y_i}}})}^T}{\Sigma _{{y_i}}}({{\rm{w}}_C} - {{\rm{w}}_{{y_i}}})}
\end{array}} \right].
\end{aligned}
\end{equation}

Obviously, the compensation is category-level and determined with prior knowledge. In addition, the compensation direction is negative as the loss is increased for each training sample. The term is heavily dependent on the co-variance matrix $\Sigma_{y_i}$, which can be further optimized via meta learning by minimizing the following loss on a validation set $\Omega$: 

\begin{equation}
{\Sigma ^*} = \arg \mathop {\min }\limits_\Sigma  \sum\limits_{j \in \Omega } {l\left( {\mathbb{S}\left( {{u_j}} \right),{y_j};{\Theta ^*}(\Sigma)} \right)},
\end{equation}
which is just the meta implicit data augmentation (MetaSAug) proposed by Li et al.~\cite{li2021metasaug}. MetaSAug is quite effective in long-tail classification.

(8) {\bf Arcface~\cite{liu2019adaptiveface}.}
Arcface is a classical face recognition loss defined as follows:
\begin{equation}
\small
\mathcal{L}{\rm{ =  - }}\sum\nolimits_{i } {\frac{\exp [s_i(cos(\theta_{i,y_i}+m))]}{{{\exp [s_i(cos(\theta_{i,y_i}+m))]}}+\sum\nolimits_{c \neq y_i} \exp [s_i(cos(\theta_{i,c}))]}},
\end{equation}
where $\theta_{i,c}$ is the angle between the weight $w_{c}$ and the feature $\tilde{x}_i$ which are defined in the description for ISDA; $m$ is a hyper-parameter. Indeed, $m$ does not belong to the five compensation targets in our taxonomy. It is a corpus-level term and determined via hyper-parameter tuning.

Wang et al.~\cite{wang2021meta} proposed a new Arcface loss, namely, Balancedloss, with the category-level compensation. The loss is defined as 
\begin{equation}
\small
\mathcal{L}{\rm{ =  - }}\sum\nolimits_{i } {\frac{\exp [s_i(cos(\theta_{i,y_i}+m_{g_i}))]}{{{\exp [s_i(cos(\theta_{i,y_i}+m_{g_i}))]}}+\sum\nolimits_{c \neq y_i} \exp [s_i(cos(\theta_{i,c}))]}},
\end{equation}
where ${g_i}$ is the skin-tone category of the $j$-th sample. Obviously, $m_{g_i}$ is a category-level term. It can be optimized via meta learning:
\begin{equation}
\small
m_g^*=arg\underset{\{m_{g_j}\}}{\min} \sum_{j \in \Omega} l(\Theta(m_{g_j})),
\end{equation}
which is proven to be quite effective in the experiments conducted by Wang et al.~\cite{wang2021meta}.

Other numerous typical methods such as Robust nonrigid ICP~\cite{hontani2012robust}, D2L~\cite{ma2018dimensionality}, DAC~\cite{thulasidasan2019combating}, Deep self-learning~\cite{han2019deep}, LDAM~\cite{cao2019learning}, MRFL~\cite{zhao2020multi}, Robust regression~\cite{slawski2019linear}, and Bootstrapping loss~\cite{reed2014training} can also be explained with compensation learning.

\section{Two New Learning Method Examples}
This section introduces two method examples by introducing the idea of compensation learning into existing algorithms.
\subsection{\textit{l}1-based Logit Compensation}
An example is given to explain how logit compensation works. Assume that the inferred logit vector of a noisy sample $x_i$ is $u_i = [3.0, 0.8, 0.2]$ and its (noisy) label $y_i$ is [0, 1, 0]. The cross-entropy loss incurred by this training sample is 2.36. This loss negatively affects in training because $y_i$ is noisy. To reduce the negative influence, if a compensation vector (e.g., [-1, 2, 0]) is learned, then the new logit vector becomes [2, 2.8, 0.2]. Consequently, the cross-entropy loss of $x_i$ is 0.42, which is much lower than 2.36. When $l1$-norm is used, the training loss is
\begin{equation}\label{p27}
\mathcal{L}  = \sum\nolimits_i {l(\mathbb{S}({u_i} + {v _i}),{y_i})}  + \lambda {\rm{|}}{v _i}{\rm{|}},
\end{equation}
where $v_i$ is the logit compensation vector and it is trainable during the training stage. If no noisy and quite hard samples are present, $v_i$ will approach to zero for all training samples. This method is called \textit{LogComp} for brevity. The detailed steps are described in Algorithm 1.

\subsection{Mixed Positive and Negative Compensation}
We observed that large compensations (i.e., $v_i$) concentrate in samples with large losses during the running of LogComp in the experiments. Let $l_i=l(\mathbb{S}(u_i),y_i)$. Motivated by adversarial training, ~(\ref{p27}) is modified into the following form
\begin{equation}\label{p28}
\mathcal{L}  =\sum\limits_{i:{l_i} \ge \tau } {\mathop {\min }\limits_{\left\| {{v _i}} \right\| \le \epsilon } } l\left( {\mathbb{S}\left( {{u_i} + {v _i}} \right),{y_i}} \right) + \sum\limits_{i:{l_i} \textless \tau} {{l_i}} .
\end{equation}
Compared with~(\ref{p27}),~(\ref{p28}) has one more hyper-parameter. Nevertheless,~(\ref{p28}) is more flexible than~(\ref{p27}). The results on image classification show that~(\ref{p28}) is better than~(\ref{p27}) if appropriate $\tau$ and $\epsilon$ are used.

Further, negative feature compensation can be used to increase the influences of samples whose losses are below the threshold $\tau$ in the optimization. A mixed compensation is subsequently obtained with the following loss:
\begin{equation}\label{p29}
\begin{aligned}
\mathcal{L}  &=\sum\limits_{i:{l_i} \ge \tau } {\mathop {\min }\limits_{\left\| {{v _i}} \right\| \le \epsilon_1 } } l\left(\mathbb{S} ({{{u_i} + {v _i}}),{y_i}} \right) 
\\
&+ \sum\limits_{i:{l_i } \textless \tau} {\mathop {\max }\limits_{\left\| {{\delta _i}} \right\| \le \epsilon_2 } } l\left( {\mathbb{S}  (f\left( {{x_i} + {\delta _i}}) \right),{y_i}} \right).
\end{aligned}
\end{equation}

The main difference between~(\ref{p29}) and the adversarial training loss~\cite{madry2017towards} is that the losses of quite hard (including noisy) samples are not increased any more in~(\ref{p29}). Instead, the losses of these samples are reduced as in~(\ref{p29}).
When $\tau >  \max \limits_i\{l_i\}$, only the maximization part exists and the whole loss becomes the adversarial training loss; when $\epsilon_2 =0$,~(\ref{p29}) is reduced to~(\ref{p28}).

The minimization part in both~(\ref{p28}) and~(\ref{p29}) can be solved with an optimization approach similar to PGD \cite{madry2017towards}. This method is called \textit{MixComp} for brevity. The PGD-like optimization for the minimization part in Eqs. (44) and (45) is as follows. First, we have
\begin{equation}
\frac{{\partial l(\mathbb{S}({u_i} + {v_i}),{y_i})}}{{\partial {v_i}}}\left| {_{{v_i} = 0}} \right. = \mathbb{S}({u_i}) - {\hat y_i},
\end{equation}
where $\hat y_i$ is the one-hot vector of $y_i$. Therefore, $v_i$ can be calculated by
\begin{equation}\label{app:a17}
{v_i} = \eta ({\hat y_i}-\mathbb{S}({u_i})),
\end{equation}
where $\eta$ is the hyper-parameter. Accordingly, the updating of $u_i$ is
\begin{equation}
{u'_i} = {u_i}{\rm{ + }}\eta ({\hat y_i}-\mathbb{S}({u_i})).
\end{equation}

In our implementation, only one updating step is used. Consequently, if $\infty$-norm is used, then we have
\begin{equation}
{\rm{|}}{v_i}{\rm{| = |}}\eta (\mathbb{S}({u_i}) - {\hat y_i})| \le |\eta ||(\mathbb{S}({u_i}) - {\hat y_i})| \le \eta.
\end{equation}
Therefore, we use $\eta$ to control the bound (i.e., $\epsilon_1$) of $v_i$. The detailed steps of MixComp are described in Algorithm 2.

\renewcommand{\algorithmicrequire}{\textbf{Input:}} 
\renewcommand{\algorithmicensure}{\textbf{Output:}} 
\begin{algorithm}[H] 
\caption{LogComp} 
\label{alg::1} 
\begin{algorithmic}[1] 
\Require 
Training set $S = \{ {x_i},{\rm{ }}{y_i}\}$, $i = 1, \cdots, N$; 
hyper-parameters $\lambda$; 
\#Epoch; 
\#Batch;
and learning rate.
\Ensure 
Model $f({x,{\rm{ }}{\mathop{\rm \textbf{w}}\nolimits} })$. 
\State \textbf{Initialization:} $v=\textbf{0}$ for each training sample, $\textbf{w}$ as ${\textbf{w}^{\left( 0 \right)}}$; 
\Repeat 
\State $t$ = $1,\cdots,$ \#Epoch
\State \quad$k$ = $1,\cdots,$ \#Batch
\State \quad\quad Generate mini-batch $D_k$ from $S$; 
\State \quad\quad Calculate loss based on Eq.~(\ref{p27}); 
\State \quad\quad Update $\textbf{w}$ via SGD;
\Until stable accuracy in the validation set. 
\end{algorithmic} 
\end{algorithm}

\begin{algorithm}[H] 
\caption{MixComp}\label{alg::2} 
\begin{algorithmic}[1] 
\Require 
Training set $S = \{ {x_i},{\rm{ }}{y_i}\}$, $i = 1, \cdots, N$; 
\#Epoch; 
\#Batch;
learning rate; $\epsilon_1$; $\epsilon_2$;
and $\tau$.
\Ensure 
Model $f({x,{\rm{ }}{\mathop{\rm \textbf{w}}\nolimits} })$. 
\State \textbf{Initialization:} $\textbf{w}$ as ${\textbf{w}^{\left( 0 \right)}}$; 
\Repeat 
\State $t$ = $1,\cdots,$ \#Epoch
\State \quad$k$ = $1,\cdots,$ \#Batch
\State \quad\quad Generate mini-batch $D_k$ from $S$;
\State \quad\quad Infer $v_i$ according to Eq.~(\ref{app:a17}) for samples with a lower loss than $\tau$;  
\State \quad\quad Infer $\delta_i$ for the rest samples according to PGD optimization;  
\State \quad\quad Calculate loss based on Eq.~(\ref{p29}); 
\State \quad\quad Update $\textbf{w}$ using SGD;
\Until stable accuracy in the validation set. 
\end{algorithmic} 
\end{algorithm}

\section{Experiments}
This section evaluates our methods (LogComp and MixComp) in image classification and text sentiment analysis when there are noisy labels.

\subsection{Competing Methods}
As our proposed methods belong to the end-to-end noise-aware solution, the following methods are compared: soft/hard Bootstrapping~\cite{reed2014training}, label smoothing~\cite{szegedy2016rethinking}, online label smoothing~\cite{zhang2021delving}, progressive self label correction (ProSelfLC)~\cite{wang2021proselflc}, and PGD-based adversarial training (PGD-AT)~\cite{madry2017towards}. 

The parameter settings are detailed in the corresponding subsections. In all experiments, the average classification accuracy and standard deviation of three repeated runs are recorded for each comparison. 
\subsection{Image Classification}\label{sec:c10-1}
\begin{table*}[!htb]
\caption{Classification accuracies(\%) on CIFAR-10.}
\centering
\begin{tabular}{c|c|ccc|ccc}
\hline
& & \multicolumn{3}{c|}{Random noise} & \multicolumn{3}{c}{Pair noise}          \\ \hline
& 0\%           & 10\%          & 20\%      & 30\%      & 10\%     & 20\%     & 30\%      \\ \hline
Base (ResNet-20)       & 91.79$\pm$0.31       & 88.78$\pm$0.33       & 87.55$\pm$0.32   & 85.85$\pm$0.37   & 90.32$\pm$0.19  & 89.28$\pm$0.14  & 87.06$\pm$0.23   \\
Soft Bootstrapping     & 91.83$\pm$0.12       & 89.37$\pm$0.18       & 87.52$\pm$0.37   & 85.59$\pm$0.33   & 90.44$\pm$0.23  & 89.16$\pm$0.22  & 87.08$\pm$0.25   \\
Hard Bootstrapping     & 92.06$\pm$0.10       & 89.61$\pm$0.20       & 88.07$\pm$0.32   & 86.37$\pm$0.26   & 90.34$\pm$0.18  & 89.54$\pm$0.25  & 86.86$\pm$0.19   \\
Label Smoothing        & 92.12$\pm$0.14       & 90.15$\pm$0.09       & 88.54$\pm$0.18   & 86.82$\pm$0.16   & 90.63$\pm$0.22  & 90.12$\pm$0.06  & 88.28$\pm$0.42   \\
Online Label Smoothing & 92.18$\pm$0.15       & 89.84$\pm$0.14       & 88.19$\pm$0.15   & 86.08$\pm$0.22   & 90.65$\pm$0.18  & 89.52$\pm$0.08  & 87.68$\pm$0.16 \\
ProSelfLC              & 91.80$\pm$0.16       & 89.90$\pm$0.16       & 88.84$\pm$0.22   & 86.78$\pm$0.31   & 90.40$\pm$0.23  & 89.76$\pm$0.17  & 87.11$\pm$0.20 \\
PGD-AT                 & 89.90$\pm$0.08       & 87.56$\pm$0.13       & 86.87$\pm$0.13   & 84.80$\pm$0.17   & 88.90$\pm$0.15  & 88.38$\pm$0.07  & 87.44$\pm$0.10 \\
LogComp                & {\bf92.42$\pm$0.09}  & 90.99$\pm$0.06       & 90.20$\pm$0.16   & 88.81$\pm$0.18   & 91.17$\pm$0.17  & 91.13$\pm$0.08  & 89.72$\pm$0.15  \\
MixComp                & 92.26$\pm$0.04       &{\bf91.09$\pm$0.11}   & {\bf90.63$\pm$0.12} & {\bf88.98$\pm$0.15} & {\bf 91.29$\pm$0.03}  & {\bf 91.15$\pm$0.05}  & {\bf 90.01$\pm$0.16} \\ \hline
\end{tabular}
\label{tab:tabc10}
\end{table*}

\begin{table*}[!htb]
\caption{Classification accuracies(\%) on CIFAR-100.}
\centering
\begin{tabular}{c|c|ccc|ccc}
\hline
                       &         & \multicolumn{3}{c|}{Random noise} & \multicolumn{3}{c}{Pair noise} \\ \hline
                       & 0\%     & 10\%     & 20\%     & 30\%    & 10\%     & 20\%     & 30\%     \\ \hline
Base (ResNet-20)       & 67.81$\pm$0.08   & 63.67$\pm$0.29  & 60.63$\pm$0.33  & 57.82$\pm$0.35 & 63.94$\pm$0.29  & 61.22$\pm$0.03  & 55.74$\pm$0.22  \\
Soft Bootstrapping     & 68.38$\pm$0.24   & 64.01$\pm$0.23  & 60.66$\pm$0.28  & 57.97$\pm$0.23 & 64.29$\pm$0.31  & 60.71$\pm$0.23  & 56.27$\pm$0.26  \\
Hard Bootstrapping     & 67.62$\pm$0.29   & 64.28$\pm$0.33  & 60.32$\pm$0.22  & 58.09$\pm$0.19 & 63.96$\pm$0.26  & 60.69$\pm$0.29  & 56.18$\pm$0.17  \\
Label Smoothing        & 67.54$\pm$0.10   & 65.04$\pm$0.18  & 61.84$\pm$0.27  & 59.06$\pm$0.08 & 65.43$\pm$0.24  & 62.71$\pm$0.24  & 58.92$\pm$0.19 \\
Online Label Smoothing & 67.80$\pm$0.19   & 64.55$\pm$0.15  & 61.53$\pm$0.22  & 59.19$\pm$0.13 & 64.70$\pm$0.28  & 62.54$\pm$0.19 & 57.44$\pm$0.25 \\
ProSelfLC              & 68.37$\pm$0.22   & 64.64$\pm$0.28  & 62.14$\pm$0.17  & 58.93$\pm$0.24 & 65.36$\pm$0.18  & 62.57$\pm$0.16 & 59.08$\pm$0.27\\
PGD-AT                 & 64.37$\pm$0.17   & 60.39$\pm$0.24  & 57.38$\pm$0.21  & 54.23$\pm$0.16 & 60.41$\pm$0.20  & 58.08$\pm$0.13     & 54.37$\pm$0.22 \\
LogComp                &{\bf68.72$\pm$0.11} & 65.55$\pm$0.16 & 62.56$\pm$0.16 & 59.59$\pm$0.15 & 66.49$\pm$0.19 & 64.74$\pm$0.13 & 61.36$\pm$0.16  \\ 
MixComp                & 68.71$\pm$0.15   & {\bf65.79$\pm$0.14} & {\bf62.76$\pm$0.20} & {\bf60.17$\pm$0.12}& {\bf66.81$\pm$0.16} & {\bf64.83$\pm$0.11} &{\bf63.55$\pm$0.13}\\ \hline
\end{tabular}
\label{tab:tab22}
\end{table*}
Two benchmark image classification data sets, namely, CIFAR-10 and CIFAR-100~\cite{krizhevsky2009learning}, are used. CIFAR-10 contains 10 categories and CIFAR-100 contains 100 categories. The details of these two data sets are shown in~\cite{krizhevsky2009learning}. 

The synthetic label noises are simulated on the basis of the two common schemes used in~\cite{guo2018curriculumnet,han2018co,jiang2018mentornet}. The first is the random scheme in which each training sample is assigned to a uniform random label with a probability $p$. The second is the pair scheme in which each training sample is assigned to the category next to its true category on the basis of the category list with a probability $p$. The value of $p$ is set as 10\%, 20\%, and 30\%. 

The training/testing configuration used in~\cite{wang2021proselflc} is followed. The parameter settings are as follows. The batch size and learning rate are set as 128 and 0.1, respectively. Other parameter settings are detailed in Appendix~\ref{app:9}.

\begin{figure}[ht]
    \centering
    \includegraphics[width=0.25\linewidth]{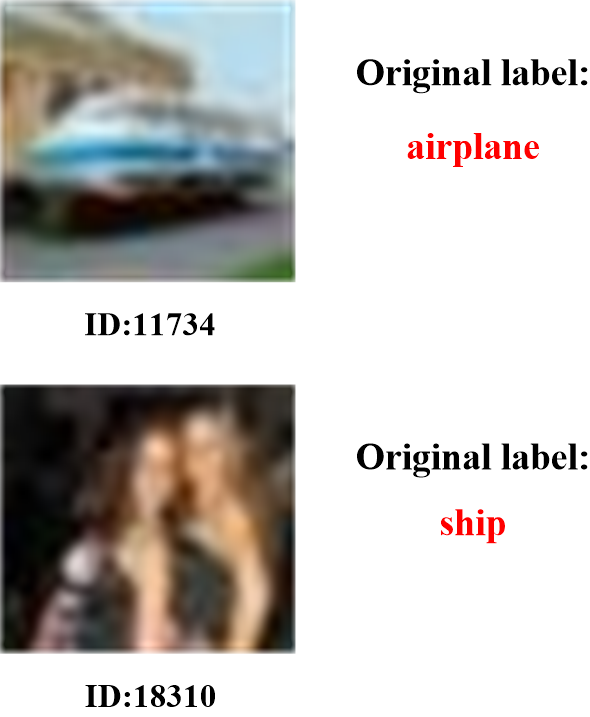}
    \caption{Samples with higher \textit{l}1-norm values of logit compensation whose labels seem erroneous.}
    \label{fig:figa215}
\end{figure}

The results are shown in Tables~\ref{tab:tabc10} and~\ref{tab:tab22}, respectively, when ResNet-20~\cite{he2016deep} is used as the base neural network. Our methods, MixComp and LogComp, achieve twelve and two highest accuracies among the fourteen comparisons, respectively. The results of MixComp are obtained when $\epsilon_2$ equals to 0, indicating that only positive compensation is useful for the (clean) accuracy when there are noises. Indeed, both the hyper-parameters $\epsilon_2$ and $\tau$ balance the trade-off between positive and negative compensations. 

\begin{table*}[!ht]
\caption{An ablation study of MixComp on CIFAR-10 (\%).}
\centering
\begin{tabular}{c|ccccc}
\hline
Random noise                         & 0\%                 & 10\%               & 20\%                & 30\% \\ \hline
Baseline (ResNet-20)                 & 91.79$\pm$0.31       & 88.78$\pm$0.33       & 87.55$\pm$0.32   & 85.85$\pm$0.37  \\
Only pos. comp. ($\epsilon_2 =0$)    & {\bf92.26$\pm$0.04}  &{\bf91.09$\pm$0.11}   & {\bf90.63$\pm$0.12} & {\bf88.98$\pm$0.15}\\
Only neg. comp. ($\epsilon_1 =0$)    & 91.66$\pm$0.12       & 88.69$\pm$0.22     & 87.33$\pm$0.10      & 85.71$\pm$0.31  \\
Both directions                      & 91.91$\pm$0.19       & 90.11$\pm$0.27     & 89.75$\pm$0.17      & 88.17$\pm$0.16   \\
\hline
\end{tabular}
\label{tab:tabnew3}
\end{table*}

\begin{table*}[ht]
\centering
\caption{Performance variations under different values of $\epsilon_2$.}
\begin{tabular}{c|ccccc}
\hline
                         & 0 & 2/255 & 4/255  & 6/255 & 8/255\\ \hline
Clean accuracy(\%)       &91.09$\pm$0.11    &90.11$\pm$0.27    & 89.77$\pm$0.18    &88.67$\pm$0.15   &88.30$\pm$0.19  \\
Adversarial accuracy(\%) &11.57$\pm$0.36    &53.38$\pm$0.31     & 64.95$\pm$0.24    &68.20$\pm$0.16   &70.25$\pm$0.14 \\ \hline
\end{tabular}
\label{tab:tab5}
\end{table*}

\begin{table*}[!htb]
\caption{Classification accuracies(\%) on CIFAR-10 (0\% noise).}
\centering
\begin{tabular}{c|c|c|c|c}
\hline
                       & ResNet-32        & ResNet-44           & ResNet-56           & ResNet-110       \\ \hline
Base                   & 92.50$\pm$0.26   & 92.82$\pm$0.15      & 93.03$\pm$0.34      & 93.51$\pm$0.18  \\
Soft Bootstrapping     & 92.40$\pm$0.17   & 92.83$\pm$0.16      & 93.43$\pm$0.27      & 94.08$\pm$0.29 \\
Hard Bootstrapping     & 92.19$\pm$0.23   & 92.94$\pm$0.11      & 93.38$\pm$0.25      & 94.02$\pm$0.23\\
Label Smoothing        & 92.75$\pm$0.24   & 92.89$\pm$0.18      & 93.05$\pm$0.23      & 93.92$\pm$0.43 \\
Online Label Smoothing & 92.61$\pm$0.19   & 92.93$\pm$0.34      & 93.41$\pm$0.20      & 93.54$\pm$0.18  \\
ProSelfLC              & 92.87$\pm$0.22   & 92.98$\pm$0.28      & 93.21$\pm$0.19      & 93.58$\pm$0.37\\
PGD-AT                 & 90.66$\pm$0.16   & 91.31$\pm$0.19      & 91.80$\pm$0.22      & 91.98$\pm$0.15 \\
LogComp                &{\bf93.42$\pm$0.11}&{\bf93.59$\pm$0.09} &{\bf93.80$\pm$0.17}  &{\bf94.40$\pm$0.12} \\ 
MixComp                & 93.00$\pm$0.15   &93.18$\pm$0.13       &93.38$\pm$0.21       & 94.35$\pm$0.10 \\ \hline
\end{tabular}
\label{tab:tabadd1}
\end{table*}

\begin{table*}[!htb]
\caption{Classification accuracies(\%) on CIFAR-100 (0\% noise).}
\centering
\begin{tabular}{c|c|c|c|c}
\hline
                       & ResNet-32     & ResNet-44     & ResNet-56         & ResNet-110    \\ \hline
Base                   & 69.16$\pm$0.19   & 70.02$\pm$0.19  & 70.38$\pm$0.34  & 73.18$\pm$0.12  \\
Soft Bootstrapping     & 69.76$\pm$0.25   & 70.76$\pm$0.34  & 71.01$\pm$0.40  & 74.19$\pm$0.24  \\
Hard Bootstrapping     & 69.37$\pm$0.24   & 70.06$\pm$0.29  & 70.26$\pm$0.31  & 73.35$\pm$0.18\\
Label Smoothing        & 69.91$\pm$0.27   & 70.52$\pm$0.51  & 71.49$\pm$0.29  & 74.01$\pm$0.44 \\
Online Label Smoothing & 69.53$\pm$0.22   & 70.05$\pm$0.79  & 71.06$\pm$0.26  & 73.59$\pm$0.19\\
ProSelfLC              & 69.54$\pm$0.29   & 70.39$\pm$0.35  & 70.49$\pm$0.32  & 73.42$\pm$0.24\\
PGD-AT                 & 65.94$\pm$0.18   & 66.55$\pm$0.26  & 67.58$\pm$0.29  & 70.83$\pm$0.17\\
LogComp                &{\bf71.41$\pm$0.21}&{\bf71.48$\pm$0.18} & {\bf72.73$\pm$0.24}  & {\bf75.54$\pm$0.14}\\ 
MixComp                & 70.29$\pm$0.17    &71.24$\pm$0.19 & 71.81$\pm$0.28  & 74.31$\pm$0.16 \\ \hline
\end{tabular}
\label{tab:tabadd2}
\end{table*}

\begin{table*}[!htb]
\caption{Classification accuracies(\%) on CIFAR-100 (20\% pair noise).}
\centering
\begin{tabular}{c|c|c|c|c}
\hline
                       & ResNet-32          & ResNet-44           & ResNet-56            & ResNet-110       \\ \hline
Base                   & 62.46$\pm$0.54     & 62.73$\pm$0.64      & 63.37$\pm$0.22       & 67.51$\pm$0.19  \\
Soft Bootstrapping     & 63.09$\pm$0.33     & 63.69$\pm$0.39      & 64.06$\pm$0.28       & 67.87$\pm$0.26 \\
Hard Bootstrapping     & 63.03$\pm$0.41     & 63.57$\pm$0.32      & 63.99$\pm$0.34       & 67.40$\pm$0.23\\
Label Smoothing        & 64.45$\pm$0.28     & 65.72$\pm$0.27      & 66.50$\pm$0.74       & 69.43$\pm$0.36 \\
Online Label Smoothing & 63.94$\pm$0.66     & 65.18$\pm$0.70      & 65.45$\pm$0.52       & 68.38$\pm$0.34 \\
ProSelfLC              & 64.04$\pm$0.37     & 65.04$\pm$0.44      & 63.94$\pm$0.41       & 68.86$\pm$0.26\\
PGD-AT                 & 60.13$\pm$0.31     & 60.58$\pm$0.28      & 60.02$\pm$0.32       & 65.62$\pm$0.20 \\
LogComp                & 66.50$\pm$0.27      & 66.97$\pm$0.29      & 68.57$\pm$0.25       & 71.74$\pm$0.21\\ 
MixComp                & {\bf67.14$\pm$0.23} &{\bf69.07$\pm$0.26} & {\bf68.92$\pm$0.21}  & {\bf71.86$\pm$0.18} \\ \hline
\end{tabular}
\label{tab:tabadd3}
\end{table*}


An ablation study is conducted for MixComp on CIFAR-10 (random noises) as MixComp involves both positive and negative compensations. The results in Table~\ref{tab:tabnew3} indicate that negative compensation (i.e., adversarial training) and compensation with both directions do not improve the performance yet the positive compensation achieves the best performance. Table~\ref{tab:tab5} lists the clean and adversarial accuracies of MixComp under different values of $\epsilon_2$ on the CIFAR-10 (10\% random noises). The increase of $\epsilon_2$ improves the adversarial accuracies yet reduces the clean accuracies. Although negative compensation in MixComp does not improve the clean accuracy, it benefits the adversarial accuracy.

When LogComp and MixComp are used, some original labels with
high average (positive) compensation terms are found to be erroneous. Fig.~\ref{fig:figa215} shows two samples from CIFAR-10. Their labels seem wrong. Comparisons on other base networks~\cite{he2016deep}, namely, ResNet-32, ResNet-44, ResNet-56, and ResNet-110 are also conducted. The same conclusions are still obtained. Tables~\ref{tab:tabadd1}-~\ref{tab:tabadd3} present the classification accuracies of the competing methods with the above four base networks on partial noisy rates.

\subsection{Text Sentiment Analysis}
Two benchmark data sets are used, namely, IMDB and SST-2~\cite{wang2017hybrid}. Both are binary tasks and the details can be seen in~\cite{wang2017hybrid}. Two types of label noises are added. In the first type (symmetric), the labels of the former 5\%, 10\%, and 20\% (according to there indexes in the corpus) training samples are flipped to simulate the label noises; in the second type (asymmetric), the labels of the former 5\%, 10\%, and 20\% positive samples are flipped to negative. The 300-$D$ Glove~\cite{zhao2020multi} embedding is used. The values for \#epochs, batch size, learning rate, and dropout rate follow the settings in~\cite{hong2015sentiment,wang2018sentiment}. The data split and other parameter settings are detailed in Appendix~\ref{app:10}.

The results of the competing methods on the IMDB and SST-2 for the symmetric and asymmetric label noises are shown in Tables~\ref{tab:tab1} and~\ref{tab:tab2}, respectively, when BiLSTM with attention~\cite{gers1999learning} is used as the base network. Our proposed method, MixComp, achieves the overall best results (13 highest accuracies among 14 comparisons). When no added label noises are present (0\%), both MixComp and LogComp still achieve better results than the base method BiLSTM with attention on both sets. 

\begin{table*}[!htb]
\caption{Classification accuracies(\%) on IMDB.}
\centering
\begin{tabular}{c|c|ccc|ccc}
\hline
&         & \multicolumn{3}{c|}{Symmetric noise} & \multicolumn{3}{c}{Asymmetric noise} \\ \hline
& 0\%      & 5\%       & 10\%       & 20\%       & 5\%        & 10\%       & 20\%       \\ \hline
Base (BiLSTM+attention)& 84.39$\pm$0.34    & 83.04$\pm$0.17  & 81.90$\pm$0.61        & 78.13$\pm$0.13    & 82.35$\pm$0.88    & 79.53$\pm$2.68    & 73.74$\pm$1.14    \\
Soft   Bootstrapping   & 84.79$\pm$0.87    & 83.87$\pm$0.13  & 81.11$\pm$0.62        & 79.60$\pm$1.78    & 83.36$\pm$1.11    & 80.70$\pm$2.19    & 73.52$\pm$2.65    \\
Hard   Bootstrapping   & 84.44$\pm$0.93    & 84.10$\pm$0.54  & 83.01$\pm$0.70        & 80.84$\pm$1.07    & 82.48$\pm$1.72    & 81.42$\pm$1.55    & 75.26$\pm$1.02    \\
Label Smoothing        & 84.62$\pm$0.18    & 83.14$\pm$0.24  & 82.41$\pm$0.51        & 80.73$\pm$0.20    & 82.75$\pm$0.29    & 82.28$\pm$0.33    & 74.70$\pm$0.48 \\
Online Label Smoothing & 84.83$\pm$0.51    & 84.14$\pm$0.37  & 82.09$\pm$0.54        & 80.91$\pm$1.17    & 83.78$\pm$0.77    & 81.35$\pm$0.92    & 73.75$\pm$1.38 \\
ProSelfLC              & 84.79$\pm$0.39    & 83.21$\pm$0.44  & 82.17$\pm$0.47        & 80.42$\pm$0.41    & 83.22$\pm$0.91    & 81.58$\pm$0.85    & 74.96$\pm$3.01  \\
PGD-AT                 & 85.82$\pm$0.10    & 84.12$\pm$0.37  & 83.53$\pm$0.44        & 81.48$\pm$0.18    & 82.41$\pm$0.98    & 80.75$\pm$0.73    & 72.40$\pm$2.15\\
LogComp                & 85.17$\pm$0.16    & 84.53$\pm$0.20  & 83.75$\pm$0.46        & 81.64$\pm$0.22    & 84.45$\pm$0.39    & 81.44$\pm$0.36    & 76.87$\pm$0.30\\
MixComp          & {\bf85.87$\pm$0.08}  &{\bf85.12$\pm$0.14} & {\bf84.33$\pm$0.22} & {\bf82.60$\pm$0.19}& {\bf85.12$\pm$0.18} & {\bf82.31$\pm$0.21} &{\bf77.83$\pm$0.25} \\ \hline
\end{tabular}
\label{tab:tab1}
\end{table*}

\begin{table*}[!htb]
\caption{Classification accuracies(\%) on SST-2.}
\centering
\begin{tabular}{c|c|ccc|ccc}
\hline
&         & \multicolumn{3}{c|}{Symmetric noise} & \multicolumn{3}{c}{Asymmetric noise} \\ \hline 
& 0\%       & 5\%       & 10\%      & 20\%       & 5\%         & 10\%       & 20\%       \\ \hline 
Base (BiLSTM+attention) & 83.85$\pm$0.02   & 82.71$\pm$0.05   & 81.12$\pm$0.29   & 79.72$\pm$0.03    & 82.07$\pm$0.45     & 81.46$\pm$0.19    & 79.49$\pm$0.39    \\
Soft Bootstrapping      & 83.77$\pm$0.33   & 83.25$\pm$0.17   & 82.21$\pm$0.23   & 80.40$\pm$0.42    & 82.78$\pm$0.27     & 81.66$\pm$0.44    & 79.14$\pm$0.25    \\
Hard Bootstrapping      & 83.68$\pm$0.40   & 83.18$\pm$0.22   & 81.45$\pm$0.63   & 80.50$\pm$0.16    & 82.25$\pm$0.54     & 81.73$\pm$0.23    & 79.52$\pm$0.67    \\
Label Smoothing         & 83.87$\pm$0.52   & 82.78$\pm$0.09   & 82.16$\pm$0.32   & 80.57$\pm$0.14    & 82.69$\pm$0.41     & 81.95$\pm$0.24    & 79.63$\pm$0.68 \\
Online Label Smoothing  & 83.67$\pm$0.19   & 83.34$\pm$0.14   & 82.03$\pm$0.22   & 80.61$\pm$0.39    & 82.58$\pm$0.33     & 82.20$\pm$0.32    & 79.57$\pm$0.72 \\
ProSelfLC               & 83.81$\pm$0.05   & 83.07$\pm$0.16   & 81.92$\pm$0.28   & 80.28$\pm$0.33    & 82.42$\pm$0.43     & 82.03$\pm$0.21    & 79.27$\pm$0.79 \\
PGD-AT                  & 83.88$\pm$0.14   & 82.15$\pm$0.23   & 81.81$\pm$0.19   & 80.33$\pm$0.12    & 82.36$\pm$0.23     & 81.69$\pm$0.18    & 73.81$\pm$0.24 \\
LogComp                 & 84.10$\pm$0.08   & 83.18$\pm$0.13   & 81.83$\pm$0.15   & 80.42$\pm$0.05    & {\bf82.85$\pm$0.10}& 82.23$\pm$0.15    & 78.87$\pm$0.23    \\ 
MixComp             &{\bf84.34$\pm$0.05}  &{\bf83.46$\pm$0.08}& {\bf82.31$\pm$0.16}&{\bf80.99$\pm$0.13}&82.75$\pm$0.18    &{\bf82.30$\pm$0.12}&{\bf79.80$\pm$0.21} \\ \hline
\end{tabular}
\label{tab:tab2}
\end{table*}

\begin{table*}[!ht]
\caption{An ablation study of MixComp on IMDB (\%).}
\centering
\begin{tabular}{c|ccccc}
\hline
Symmetric noise                & 0\% & 5\% & 10\%  & 20\% \\ \hline
Baseline (BiLSTM+attention)          & 84.39$\pm$0.34    & 83.04$\pm$0.17     & 81.90$\pm$0.61      & 78.13$\pm$0.13   \\
Only pos. comp. ($\epsilon_2 =0$)    & 84.65$\pm$0.11    & 83.53$\pm$0.21     & 82.91$\pm$0.33      & 80.10$\pm$0.22 \\
Only neg. comp. ($\epsilon_1 =0$)    & 85.84$\pm$0.36    & 84.87$\pm$0.22     & 83.89$\pm$0.27      & 81.73$\pm$0.24 \\
Both directions                      & {\bf85.87$\pm$0.08}  &{\bf85.12$\pm$0.14} & {\bf84.33$\pm$0.22} & {\bf82.60$\pm$0.19}  \\
\hline
\end{tabular}
\label{tab:tab3X}
\end{table*}
An ablation study is also conducted for MixComp on IMDB. Each compensation is useful and their combination achieves the best performance. The results are shown in Table~\ref{tab:tab3X}. Given that judging the sentimental states of some sentences is difficult, inevitably, some original samples are quite hard or noisy. When LogComp is used, some original labels with high average compensation terms are found to be erroneous. For example, the sentence ``\textit{Plummer steals the show without resorting to camp as nicholas' wounded and wounding uncle ralph}" is labeled as positive in the original set. More examples are listed in Tables~\ref{tab:taba1} and~\ref{tab:taba2} in the Appendix~\ref{app:8}.

\begin{figure}[htbp]
    \centering
    \includegraphics[width=0.7\linewidth]{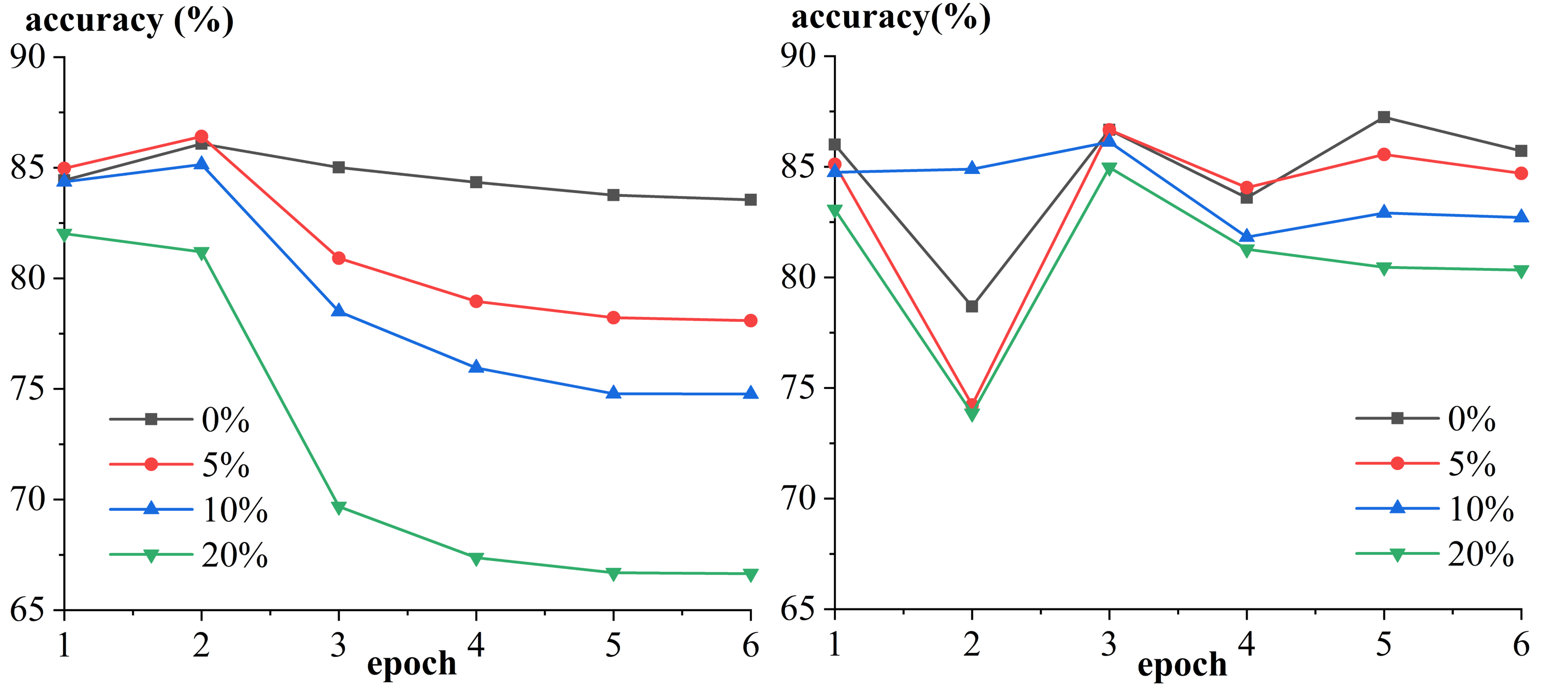}
    \caption{The validation accuracies in the first six epochs under different proportions of random noises on IMDB when using Base (left) and LogComp (right), respectively. }
    \label{fig:fig1}
\end{figure}

LogComp also achieves the second-best results on IMDB. On IMDB, the base model is usually converged in the second epoch. However, LogComp is usually converged in the third or the fifth epoch. The validation accuracies of the six epochs for the base model and our LogComp are shown in Fig.~\ref{fig:fig1}. LogComp can decelerate the convergence speed leading that the training data can be more fully trained.

\subsection{Discussion}
More extensions and new methods can be obtained based on our taxonomy. 

(1) The extension of the logit compensation described in~(\ref{p28}). As previously mentioned, each weighting method may correspond to a compensating method. Self-paced learning (SPL)~\cite{kumar2010self} is a classical sample weighting strategy in machine learning. The weights are obtained with the following objective function:
\begin{equation}
\mathop {\min }\limits_{{w_i} \in \{ 0,1\} }  \sum\nolimits_i {{w_i}l(\mathbb{S}({u_i}),{y_i}) - \lambda {w_i}}.
\end{equation}
The solution is
\begin{equation}
w_i=
\begin{cases}
1& \text{  if$\ l(\mathbb{S}(u_i),y_i) \leqslant \lambda$ } \\
0& \text{  otherwise  }
\end{cases},
\end{equation}
which indicates that the weights of samples with larger losses than $\lambda$ are set as 0. When the value of $\lambda$ is increased, more samples will participate in the model training.

\begin{figure}[ht]
    \centering
    \includegraphics[width=0.45\linewidth]{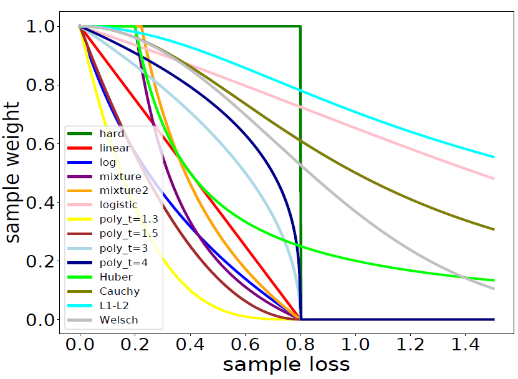}
    \caption{The curves of weights under different losses in SPL. ``Hard" represents the original SPL~\cite{wdcw}.}
    \label{fig:figa1-1}
\end{figure}

Fig.~\ref{fig:figa1-1} shows the curves of weights for the original SPL and its variants. Logit compensation can be used to implement the SPL with~(\ref{p28}) and~(\ref{27}) when the hyper-parameters $\epsilon$ and $\tau$ satisfy the following conditions:
\begin{equation}\label{27}
\tau^{t+1} > \tau^{t} \text{  and  } \epsilon > 2\max \limits_i\{||u_i||\},
\end{equation}
where $t$ is the index of the current epoch. A new method is obtained and can be called self-paced logit compensation.

\begin{figure}[ht]
    \centering
    \includegraphics[width=0.5\linewidth]{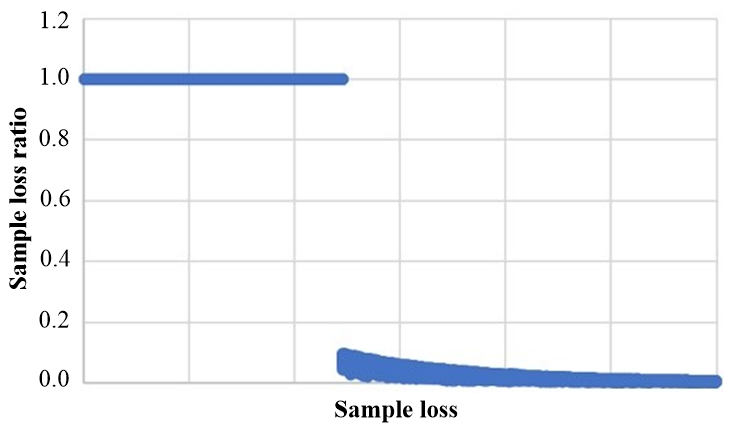}
    \caption{Loss ratio curve of self-paced logit compensation given a fixed $\eta$ and $\tau$.}
    \label{fig:figa1-2}
\end{figure}

With Eq.~(\ref{27}), similar curves can also be obtained. Fig.~\ref{fig:figa1-2} shows the curve of loss ratios (compensated loss : original loss) when $\epsilon > 2\max \limits_i\{||u_i||\}$ on the CIFAR-100 data set. The curve indicates that our strategy can also exert higher weights ($= 1$) to samples with low losses and lower weights ($\approx 0$) to samples with high losses.

(2) The extension of MixComp. Indeed, the parameters $\epsilon_1$ and $\epsilon_2$ characterize the extent of positive/negative compensations, respectively. Intuitively, an example with a larger loss should have a greater positive compensation; while an example with a lower loss should have a greater negative compensation. Therefore, the constrains for the compensation terms in (26) can be redefined as follows:
\begin{equation}
\delta _i \le \epsilon_1 [1+(l_i-\tau)/\tau]   \text{ and }
 \delta _i \le \epsilon_2 [1+(\tau-l_i)/\tau].
\end{equation}

(3) The extension of Bootstrapping. The Bootstrapping loss and the online label smoothing can be unified into the following new loss:
\begin{equation}\label{31}
\mathcal{L}  = \sum\nolimits_i {l({p_i},{y_i}{\rm{ + }}\alpha   (\beta {\widetilde{p}_{{y_i}}}{\rm{ + (1 - }}\beta {\rm{)}}{p_i} - {y_i}))},
\end{equation}
where $\widetilde{p}_{{y_i}}$ is the category-level average prediction in the previous epoch; $\alpha$ and $\beta$ are hyper-parameters and are located in [0, 1]. When $\beta$ equals 0, the above loss becomes the soft Bootstrapping loss. When $\beta$ equals 1, the loss becomes the online label smoothing loss with a little difference. Specifically, ${\tilde p}_{{y_i}}$ is defined as follows:
\begin{equation}
{\tilde p_{{y_i}}}{\rm{ = }}\frac{1}{{{Z_{{y_i}}}}}\sum\limits_{j:{y_j} = {y_i}} {({\textit{\text{conf}}}_j \times {p_j})},
\end{equation}
where ${\textit{\text{conf}}}_j$ is the prediction confidence of the prediction $p_j$, and $Z_{{y_i}}$ is the normalizer. Two typical definitions of ${\textit{\text{conf}}}_j$ are

\begin{equation}
\begin{aligned}
&\textit{\text{conf}}_j = 1\ \ or\\
&\textit{\text{conf}}_j=
\begin{cases}
1& \text{  if\ the prediction\ is\ correct  } \\
0& \text{  otherwise }
\end{cases}. 
\end{aligned}
\end{equation}

When the second definition is used and $\beta=1$, the unified loss becomes the online label smoothing. Nevertheless, the values of ${\tilde p}_{{y_i}}$ obtained by the above two definitions are close to each other when \#epoch \textgreater 5 in most data sets according to our observations. The unified new method can be called mixBootstrapping.


\section{Conclusions}
This study reveals a widely used yet under-explored machine learning strategy, namely, compensating. Machine learning methods leveraging or partially leveraging compensating comprise a new learning paradigm called compensating learning. To solidify the theoretical basis of compensation learning, a systematic taxonomy is constructed on the basis of which to compensate, the direction, how to infer, and the granularity. To demonstrate the universality of compensation learning, several existing learning methods are explained within our constructed taxonomy. Furthermore, two concrete compensation learning methods (i.e., LogComp and MixComp) are proposed. Extensive experiments suggest that our proposed methods are effective in robust learning tasks. 

\section*{Acknowledgement}
We thank Mr. Mengyang Li for his useful suggestions on the experiments.

{\small
\bibliographystyle{ieee_fullname}
\bibliography{egbib}

\begin{thebibliography}{10}\itemsep=-1pt

\bibitem{bengio2009curriculum}
Yoshua Bengio, J{\'e}r{\^o}me Louradour, Ronan Collobert, and Jason Weston.
\newblock Curriculum learning.
\newblock In {\em ICML}, pages 41--48, 2009.

\bibitem{benz2021universal}
Philipp Benz, Chaoning Zhang, Adil Karjauv, and In~So Kweon.
\newblock Universal adversarial training with class-wise perturbations.
\newblock In {\em ICME}, pages 1--6, 2021.

\bibitem{cao2019learning}
Kaidi Cao, Colin Wei, Adrien Gaidon, Nikos Arechiga, and Tengyu Ma.
\newblock Learning imbalanced datasets with label-distribution-aware margin
  loss.
\newblock In {\em NeurIPS}, pages 1565--1576, 2019.

\bibitem{chang2011ordinal}
Kuang-Yu Chang, Chu-Song Chen, and Yi-Ping Hung.
\newblock Ordinal hyperplanes ranker with cost sensitivities for age
  estimation.
\newblock In {\em CVPR}, pages 585--592, 2011.

\bibitem{cortes1995support}
Corinna Cortes and Vladimir Vapnik.
\newblock Support-vector networks.
\newblock {\em Machine learning}, 20(3):273--297, 1995.

\bibitem{Das2019A}
D. Das and Csg Lee.
\newblock A two-stage approach to few-shot learning for image recognition.
\newblock {\em IEEE Transactions on Image Processing}, 29(99):3336--3350, 2019.

\bibitem{deng2013fine}
Jia Deng, Jonathan Krause, and Li Fei-Fei.
\newblock Fine-grained crowdsourcing for fine-grained recognition.
\newblock In {\em CVPR}, pages 580--587, 2013.

\bibitem{forero2012robust}
Pedro~A Forero, Vassilis Kekatos, and Georgios~B Giannakis.
\newblock Robust clustering using outlier-sparsity regularization.
\newblock {\em IEEE Transactions on Signal Processing}, 60(8):4163--4177, 2012.

\bibitem{freund1997decision}
Yoav Freund and Robert~E Schapire.
\newblock A decision-theoretic generalization of on-line learning and an
  application to boosting.
\newblock {\em Journal of computer and system sciences}, 55(1):119--139, 1997.

\bibitem{gers1999learning}
Felix~A Gers, J{\"u}rgen Schmidhuber, and Fred Cummins.
\newblock Learning to forget: Continual prediction with lstm.
\newblock {\em Neural Computation}, 12(10):2451--2471, 2000.

\bibitem{goibert2019adversarial}
Morgane Goibert and Elvis Dohmatob.
\newblock Adversarial robustness via label-smoothing.
\newblock {\em arXiv preprint arXiv:1906.11567}, 2019.

\bibitem{goldberger2016training}
Jacob Goldberger and Ehud Ben-Reuven.
\newblock Training deep neural-networks using a noise adaptation layer.
\newblock In {\em ICLR}, 2017.

\bibitem{guo2018curriculumnet}
Sheng Guo, Weilin Huang, Haozhi Zhang, Chenfan Zhuang, Dengke Dong, Matthew~R
  Scott, and Dinglong Huang.
\newblock Curriculumnet: Weakly supervised learning from large-scale web
  images.
\newblock In {\em ECCV}, pages 139--154, 2018.

\bibitem{han2020survey}
Bo Han, Quanming Yao, Tongliang Liu, Gang Niu, Ivor~W Tsang, James~T Kwok, and
  Masashi Sugiyama.
\newblock A survey of label-noise representation learning: Past, present and
  future.
\newblock {\em arXiv preprint arXiv:2011.04406}, 2020.

\bibitem{han2018co}
Bo Han, Quanming Yao, Xingrui Yu, Gang Niu, Miao Xu, Weihua Hu, Ivor~W Tsang,
  and Masashi Sugiyama.
\newblock Co-teaching: Robust training of deep neural networks with extremely
  noisy labels.
\newblock In {\em NeurIPS}, pages 8536--8546, 2018.

\bibitem{han2019deep}
Jiangfan Han, Ping Luo, and Xiaogang Wang.
\newblock Deep self-learning from noisy labels.
\newblock In {\em ICCV}, pages 5137--5146, 2019.

\bibitem{he2016deep}
Kaiming He, Xiangyu Zhang, Shaoqing Ren, and Jian Sun.
\newblock Deep residual learning for image recognition.
\newblock In {\em CVPR}, pages 770--778, 2016.

\bibitem{hinton2015distilling}
Geoffrey Hinton, Oriol Vinyals, and Jeff Dean.
\newblock Distilling the knowledge in a neural network.
\newblock {\em arXiv preprint arXiv:1503.02531}, 2015.

\bibitem{hong2015sentiment}
James Hong and Michael Fang.
\newblock Sentiment analysis with deeply learned distributed representations of
  variable length texts.
\newblock {\em Stanford University Report}, pages 1--9, 2015.

\bibitem{hontani2012robust}
Hidekata Hontani, Takamiti Matsuno, and Yoshihide Sawada.
\newblock Robust nonrigid icp using outlier-sparsity regularization.
\newblock In {\em CVPR}, pages 174--181, 2012.

\bibitem{huang2016learning}
Chen Huang, Yining Li, Chen~Change Loy, and Xiaoou Tang.
\newblock Learning deep representation for imbalanced classification.
\newblock In {\em CVPR}, pages 5375--5384, 2016.

\bibitem{huang2019o2u}
Jinchi Huang, Lie Qu, Rongfei Jia, and Binqiang Zhao.
\newblock O2u-net: A simple noisy label detection approach for deep neural
  networks.
\newblock In {\em ICCV}, pages 3325--3333, 2019.

\bibitem{jiang2018mentornet}
Lu Jiang, Zhengyuan Zhou, Thomas Leung, Li-Jia Li, and Li Fei-Fei.
\newblock Mentornet: Learning data-driven curriculum for very deep neural
  networks on corrupted labels.
\newblock In {\em ICML}, pages 2309--2318, 2018.

\bibitem{johnson2019survey}
Justin~M Johnson and Taghi~M Khoshgoftaar.
\newblock Survey on deep learning with class imbalance.
\newblock {\em Journal of Big Data}, 6(1):1--54, 2019.

\bibitem{krizhevsky2009learning}
Alex Krizhevsky and Geoffrey Hinton.
\newblock Learning multiple layers of features from tiny images.
\newblock Technical report, 2009.

\bibitem{kumar2010self}
M Kumar, Benjamin Packer, and Daphne Koller.
\newblock Self-paced learning for latent variable models.
\newblock In {\em NeurIPS}, pages 1189--1197, 2010.

\bibitem{lee2020semantics}
Wonseok Lee, Hanbit Lee, and Sang-goo Lee.
\newblock Semantics-preserving adversarial training.
\newblock {\em arXiv preprint arXiv:2009.10978}, 2020.

\bibitem{li2019gradient}
Buyu Li, Yu Liu, and Xiaogang Wang.
\newblock Gradient harmonized single-stage detector.
\newblock In {\em AAAI}, pages 8577--8584, 2019.

\bibitem{li2019learning}
Junnan Li, Yongkang Wong, Qi Zhao, and Mohan~S Kankanhalli.
\newblock Learning to learn from noisy labeled data.
\newblock In {\em CVPR}, pages 5051--5059, 2019.

\bibitem{li2021metasaug}
Shuang Li, Kaixiong Gong, Chi~Harold Liu, Yulin Wang, Feng Qiao, and Xinjing
  Cheng.
\newblock Metasaug: Meta semantic augmentation for long-tailed visual
  recognition.
\newblock In {\em CVPR}, pages 5212--5221, 2021.

\bibitem{lin2017focal}
Tsung-Yi Lin, Priya Goyal, Ross Girshick, Kaiming He, and Piotr Doll{\'a}r.
\newblock Focal loss for dense object detection.
\newblock In {\em ICCV}, pages 2999--3007, 2017.

\bibitem{liu2019adaptiveface}
Hao Liu, Xiangyu Zhu, Zhen Lei, and Stan~Z Li.
\newblock Adaptiveface: Adaptive margin and sampling for face recognition.
\newblock In {\em CVPR}, pages 11947--11956, 2019.

\bibitem{liu2015classification}
Tongliang Liu and Dacheng Tao.
\newblock Classification with noisy labels by importance reweighting.
\newblock {\em IEEE Transactions on pattern analysis and machine intelligence},
  38(3):447--461, 2016.

\bibitem{ma2018dimensionality}
Xingjun Ma, Yisen Wang, Michael~E Houle, Shuo Zhou, Sarah Erfani, Shutao Xia,
  Sudanthi Wijewickrema, and James Bailey.
\newblock Dimensionality-driven learning with noisy labels.
\newblock In {\em ICML}, pages 3361--3370, 2018.

\bibitem{madry2017towards}
Aleksander Madry, Aleksandar Makelov, Ludwig Schmidt, Dimitris Tsipras, and
  Adrian Vladu.
\newblock Towards deep learning models resistant to adversarial attacks.
\newblock In {\em ICLR}, 2018.

\bibitem{menon2020long}
Aditya~Krishna Menon, Sadeep Jayasumana, Ankit~Singh Rawat, Himanshu Jain,
  Andreas Veit, and Sanjiv Kumar.
\newblock Long-tail learning via logit adjustment.
\newblock In {\em ICLR}, 2021.

\bibitem{Moosav2020}
Seyed-Mohsen Moosavi-Dezfooli, Omar Fawzi, Alhussein amd~Fawzi, and Pascal
  Frossard.
\newblock Universal adversarial perturbations.
\newblock In {\em CVPR}, pages 86--94, 2017.

\bibitem{natarajan2013learning}
Nagarajan Natarajan, Inderjit~S Dhillon, Pradeep Ravikumar, and Ambuj Tewari.
\newblock Learning with noisy labels.
\newblock In {\em NeurIPS}, pages 1196--1204, 2013.

\bibitem{reed2014training}
Scott Reed, Honglak Lee, Dragomir Anguelov, Christian Szegedy, Dumitru Erhan,
  and Andrew Rabinovich.
\newblock Training deep neural networks on noisy labels with bootstrapping.
\newblock In {\em ICLR}, 2015.

\bibitem{ren2018learning}
Mengye Ren, Wenyuan Zeng, Bin Yang, and Raquel Urtasun.
\newblock Learning to reweight examples for robust deep learning.
\newblock In {\em ICML}, pages 4331--4340, 2018.

\bibitem{shafahi2020universal}
Ali Shafahi, Mahyar Najibi, Zheng Xu, John Dickerson, Larry~S Davis, and Tom
  Goldstein.
\newblock Universal adversarial training.
\newblock In {\em AAAI}, pages 5636--5643, 2020.

\bibitem{mwn2019}
Jun Shu, Qi Xie, Lixuan Yi, Qian Zhao, Sanping Zhou, Zongben Xu, and Deyu Meng.
\newblock Meta-weight-net: Learning an explicit mapping for sample weighting.
\newblock In {\em NeurIPS}, pages 1917--1928, 2019.

\bibitem{slawski2019linear}
Martin Slawski, Emanuel Ben-David, et~al.
\newblock Linear regression with sparsely permuted data.
\newblock {\em Electronic Journal of Statistics}, 13(1):1--36, 2019.

\bibitem{song2020learning}
Hwanjun Song, Minseok Kim, Dongmin Park, Yooju Shin, and Jae-Gil Lee.
\newblock Learning from noisy labels with deep neural networks: A survey.
\newblock {\em arXiv preprint arXiv:2007.08199}, 2020.

\bibitem{szegedy2016rethinking}
Christian Szegedy, Vincent Vanhoucke, Sergey Ioffe, Jon Shlens, and Zbigniew
  Wojna.
\newblock Rethinking the inception architecture for computer vision.
\newblock In {\em CVPR}, pages 2818--2826, 2016.

\bibitem{thulasidasan2019combating}
Sunil Thulasidasan, Tanmoy Bhattacharya, Jeff Bilmes, Gopinath Chennupati, and
  Jamal Mohd-Yusof.
\newblock Combating label noise in deep learning using abstention.
\newblock In {\em ICML}, pages 6234--6243, 2019.

\bibitem{wang2017hybrid}
Chenglong Wang, Feijun Jiang, and Hongxia Yang.
\newblock A hybrid framework for text modeling with convolutional rnn.
\newblock In {\em KDD}, pages 2061--2069, 2017.

\bibitem{wang2021meta}
Mei Wang, Yaobin Zhang, and Weihong Deng.
\newblock Meta balanced network for fair face recognition.
\newblock {\em IEEE Transactions on Pattern Analysis and Machine Intelligence},
  2021.

\bibitem{wang2021proselflc}
Xinshao Wang, Yang Hua, Elyor Kodirov, David~A Clifton, and Neil~M Robertson.
\newblock Proselflc: Progressive self label correction for training robust deep
  neural networks.
\newblock In {\em CVPR}, pages 752--761, 2021.

\bibitem{wang2017robust}
Yixin Wang, Alp Kucukelbir, and David~M Blei.
\newblock Robust probabilistic modeling with bayesian data reweighting.
\newblock In {\em ICML}, pages 3646--3655, 2017.

\bibitem{wang2019implicit}
Yulin Wang, Xuran Pan, Shiji Song, Hong Zhang, Cheng Wu, and Gao Huang.
\newblock Implicit semantic data augmentation for deep networks.
\newblock In {\em NeurIPS}, pages 12614--12623, 2019.

\bibitem{wang2018sentiment}
Yequan Wang, Aixin Sun, Jialong Han, Ying Liu, and Xiaoyan Zhu.
\newblock Sentiment analysis by capsules.
\newblock In {\em WWW}, pages 1165--1174, 2018.

\bibitem{wang2020generalizing}
Yaqing Wang, Quanming Yao, James~T Kwok, and Lionel~M Ni.
\newblock Generalizing from a few examples: A survey on few-shot learning.
\newblock {\em ACM Computing Surveys (CSUR)}, 53(3):1--34, 2020.

\bibitem{wang2020training}
Zhen Wang, Guosheng Hu, and Qinghua Hu.
\newblock Training noise-robust deep neural networks via meta-learning.
\newblock In {\em CVPR}, pages 4523--4532, 2020.

\bibitem{wdcw}
Xiaoxia Wu, Ethan Dyer, and Behnam Neyshabur.
\newblock When do curricula work?
\newblock In {\em ICLR}, 2021.

\bibitem{Yao2019Deep}
J. Yao, J. Wang, I.~W. Tsang, Y. Zhang, J. Sun, C. Zhang, and R. Zhang.
\newblock Deep learning from noisy image labels with quality embedding.
\newblock {\em IEEE Transactions on Image Processing}, 28:1909--1922, 2019.

\bibitem{zhang2021delving}
Chang-Bin Zhang, Peng-Tao Jiang, Qibin Hou, Yunchao Wei, Qi Han, Zhen Li, and
  Ming-Ming Cheng.
\newblock Delving deep into label smoothing.
\newblock {\em IEEE Transactions on Image Processing}, 30:5984--5996, 2021.

\bibitem{zhang2020adversarial}
Wei~Emma Zhang, Quan~Z Sheng, Ahoud Alhazmi, and Chenliang Li.
\newblock Adversarial attacks on deep-learning models in natural language
  processing: A survey.
\newblock {\em ACM Transactions on Intelligent Systems and Technology (TIST)},
  11(3):1--41, 2020.

\bibitem{zhao2020multi}
Liang Zhao, Tianyang Zhao, Tingting Sun, Zhuo Liu, and Zhikui Chen.
\newblock Multi-view robust feature learning for data clustering.
\newblock {\em IEEE Signal Processing Letters}, 27:1750--1754, 2020.

\end{thebibliography}
}

\appendix
\setcounter{equation}{0}
\setcounter{table}{0}
\setcounter{figure}{0}
\renewcommand\theequation{a-\arabic{equation}}
\renewcommand\thetable{A-\arabic{table}}
\renewcommand\thefigure{A-\arabic{figure}}

\section{Meta Logit Adjustment}\label{app:4}
In Eq.~(\ref{pp26}) of the paper, the hyper-parameter $\tau$ is fixed for all categories. A category-wise setting for $\tau$ may be useful. Therefore, a new logit adjustment with meta optimization on $\tau$ is proposed and called Meta logit adjustment.
Let $\Omega$ be the validation set for meta optimization. According to Eqs. (\ref{9}--\ref{11}) in the paper, the new loss is

\begin{equation}\label{aa-7}
\mathcal{L} = {\rm{ - }}\sum\limits_{{x_i} \in S} {\log \frac{{{e^{{u_{i,{y_i}}} + {\tau _{{y_i}}}\log {\pi _{{y_i}}}}}}}{{\sum\nolimits_y {{e^{{u_{i,y}} + {\tau _{{y_i}}}\log {\pi _y}}}} }}}.
\end{equation}

Given a value for $\tau = \{\tau _1, …, \tau_C\}$, the network parameter $\Theta$ can be obtained by solving
\begin{equation}\label{aa-8}
{\Theta ^{\rm{*}}}(\tau ) = \arg \mathop {\min }\limits_\Theta  {\rm{ - }}\sum\limits_{{x_i} \in S} {\log \frac{{{e^{{u_{i,{y_i}}} + {\tau _{{y_i}}}\log {\pi _{{y_i}}}}}}}{{\sum\nolimits_y {{e^{{u_{i,y}} + {\tau _{{y_i}}}\log {\pi _y}}}} }}}.
\end{equation}
After ${\Theta ^{\rm{*}}}(\tau )$  is obtained, $\tau$ can be optimized by solving
\begin{equation}\label{aa-9}
{\tau ^{\rm{*}}} = \arg \mathop {\min }\limits_\tau  {\rm{ - }}\sum\limits_{{x_i} \in \Omega } {l\left( {\text{softmax} \left( {f\left( {{x_i}:{\Theta ^ * }\left( \tau  \right)} \right),{y_i}} \right)} \right)}.
\end{equation}

Eqs.~(\ref{aa-8}) and~(\ref{aa-9}) are solved alternatively. The detailed optimization steps are similar to those used in MetaSDA~\cite{li2021metasaug}, Meta-Weight-Net~\cite{mwn2019}, and other meta optimization studies.

\section{Parameter setting in image classification}\label{app:9}

For the two data sets, the \#epochs are set as 300. The $\lambda$  in LogComp is searched in \{0.25, 0.35\} and the learning rate for the compensation variable in LogComp is searched in \{1.5, 3, 4.5, 6\}. In MixComp, $\epsilon_1 (i.e.,  \eta)$ is searched in \{0.5, 1.5, 2, 3, 4, 5\}, and $\epsilon_2$ is searched in \{0, 8/255, 10/255, 12/255\}. $\tau$ is determined  according to the top-$pro$ percent of ordered losses, and the value of $pro$ is searched in \{0, 5, 7.5, 15, 25, 35, 45, 50\}. In Soft/Hard Bootstrapping, Label Smoothing, and online label smoothing, the parameters follow the settings in~\cite{zhang2021delving}. In ProSelfLC, the parameters follow the settings in~\cite{wang2021proselflc}. In PGD-AT, $\epsilon_2$ is is searched in \{8/255, 10/255, 12/255\}.
\section{Parameter setting in text sentiment analysis}\label{app:10}
\vspace{-0.1cm}
For IMDB data set, the batch size is set as 64; the learning rate is set as 0.001; the number of epochs is set as 6; the proportion of train/val/test data is 4:1:5; the embedding dropout is set as 0.5; the dimension of hidden vectors is 100. In LogComp, the learning rate for the compensation variable is searched in \{0.6, 0.7, 0.8, 0.9, 1\}, and the $\lambda$ is searched in \{0.75, 1\}. In MixComp, $\epsilon_1 (i.e.,  \eta)$ is searched in \{0, 0.075, 0.15, 0.25, 0.5, 0.75, 1\}, and $\epsilon_2$ is searched in \{0, 0.005, 0.01, 0.015\}. $\tau$ is determined  according to the top-$pro$ percent of ordered losses, and the value of $pro$ is searched in \{0, 5, 7.5, 15, 25, 35, 45, 50\}.

For SST-2 data set, the batch size is set as 32; the learning rate is set as 0.0001; the number of epochs is set as 50; the division of train/val/test data follows the default split; the embedding dropout is set as 0.7; the dimension of hidden vectors is 256. In LogComp, the learning rate for the compensation variable is searched in \{0.02, 0.025, 0.03, 0.035, 0.04\}, and the $\lambda$ is searched in \{0.75, 1\}. In MixComp, $\epsilon_1 (i.e.,  \eta)$ is searched in \{0, 0.075, 0.15, 0.25, 0.5, 0.75, 1\}, and $\epsilon_2$ is searched in \{0, 0.005, 0.01, 0.015\}. $\tau$ is determined  according to the top-$pro$ percent of ordered losses, and the value of $pro$ is searched in \{0, 5, 7.5, 15, 25, 35, 45, 50\}.

For the two data sets, in Soft/Hard Bootstrapping, Label Smoothing, and online label smoothing, the parameters follow the settings in~\cite{zhang2021delving}. In ProSelfLC, the parameters follow the settings in~\cite{wang2021proselflc}. In PGD-AT, $\epsilon_2$ is searched in \{0.005, 0.01, 0.015\}.
\vspace{-0.3cm}
\section{More sentence examples}\label{app:8}
\vspace{-0.1cm}
Table~\ref{tab:taba1} shows some samples with higher $l1$-norm values of logit compensation whose labels are erroneous. Without contexts, we believe that these labels are wrong. Some readers may consider that the labels are correct in certain contexts. In our view, it is inappropriate to assume that annotators are familiar with these contexts in advance. Table~~\ref{tab:taba2} shows some samples that are difficult to predict by machines. Their $l1$-norm values are also high.

\begin{table}[h]
\centering
\caption{Sentences with wrong labels.}
\begin{tabular}{l|c|c}
\hline
\multicolumn{1}{c|}{Sample} & Original label & Our label \\ \hline
\multicolumn{1}{m{10cm}|}{the exploitative, clumsily staged violence overshadows everything, including most of the actors.}     & \multicolumn{1}{m{2cm}<\centering|}{1}              & 0         \\ \hline
\multicolumn{1}{m{10cm}|}{.. a fascinating curiosity piece -- fascinating, that is, for about ten minutes.}      & 0              & 1         \\ \hline
\multicolumn{1}{m{10cm}|}{this is a great movie. I love the series on tv and so I loved the movie. One of the best things in the movie is that Helga finally admits her deepest darkest secret to Arnold!!! that was great. i loved it it was pretty funny too. It's a great movie! Doy!!!}      & 0              & 1         \\ \hline
\end{tabular}
\vspace{-0.15in}
\label{tab:taba1}
\end{table}

\begin{table}[h]
\centering
\caption{Sentences that are difficult to predict by machines.}
\begin{tabular}{l|c}
\hline
\multicolumn{1}{c|}{Sample} & Original label\\ \hline
\multicolumn{1}{m{10cm}|}{it 's a boring movie about a boring man, made watchable by a bravura performance from a consummate actor incapable of being boring.}     & 1              \\ \hline
\multicolumn{1}{m{10cm}|}{she is a lioness, protecting her cub, and he a reluctant villain, incapable of controlling his crew.}      & 1              \\ \hline
\multicolumn{1}{m{10cm}|}{it made me want to get made-up and go see this movie with my sisters.}      & 1            \\ \hline
\end{tabular}
\label{tab:taba2}
\end{table}

\begin{figure}[h]
    \centering
    \includegraphics[width=0.4\linewidth]{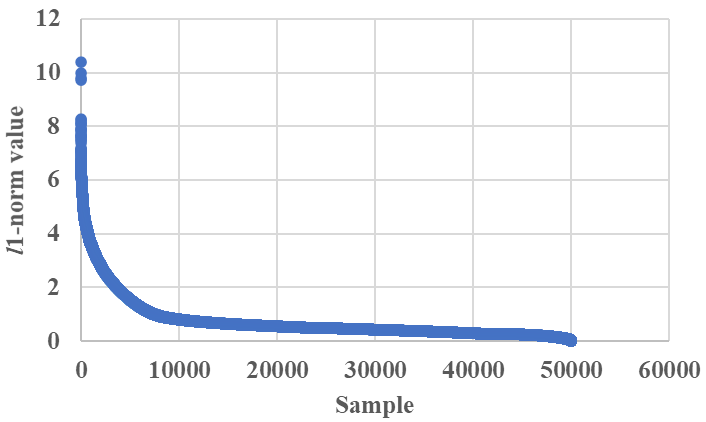}
    \caption{Distribution of $l1$-norm of compensated logit vectors on CIFAR-10.}
    \label{fig:figa213}
\end{figure}

\begin{figure}[h]
    \centering
    \includegraphics[width=0.4\linewidth]{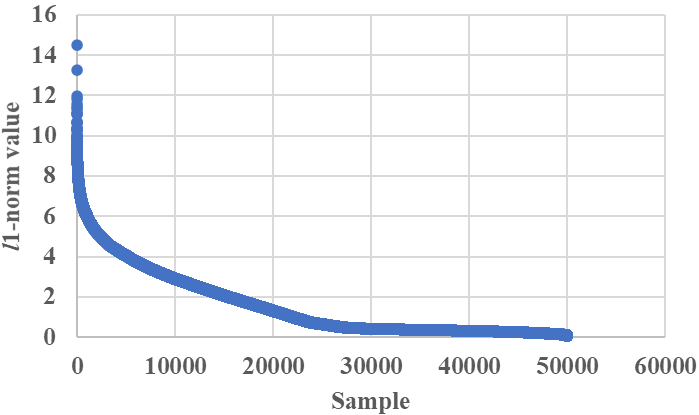}
    \caption{Distribution of $l1$-norm of compensated logit vectors on CIFAR-100.}
    \label{fig:figa214}
\end{figure}

In addition, we plot the distribution of $l1$-norm of compensated logit vectors when using LogComp on CIFAR-10 and CIFAR-100 data sets when no added label noises are present (0\%). The results are shown in Fig.~\ref{fig:figa213} and Fig.~\ref{fig:figa214}. Both distribution curves show a long-tail trend, which is quite reasonable.


\end{document}